\documentclass[pdflatex,sn-mathphys-num]{sn-jnl}% Math and Physical Sciences Numbered Reference Style

\newif\ifpreprint
\preprinttrue % Set to \preprinttrue for preprint style, \preprintfalse for submission version

\ifpreprint
\usepackage{arxiv}
\fi

\usepackage{graphicx}%
\usepackage{multirow}%
\usepackage{amsmath,amssymb,amsfonts}%
\usepackage{amsthm}%
\usepackage{mathrsfs}%
\usepackage[title]{appendix}%
\usepackage{xcolor}%
\usepackage{transparent}%
\usepackage{textcomp}%
\usepackage{manyfoot}%
\usepackage{booktabs}%
\usepackage{algorithm}%
\usepackage{algorithmicx}%
\usepackage{algpseudocode}%
\usepackage{listings}%
\usepackage{subcaption}%

\usepackage{xfrac}

\usepackage{tikz}
\usepackage{tikz-3dplot}
\usepackage{tkz-euclide}
\usepackage{pgfplots}
\usepackage{pgfplotstable}
\pgfplotsset{compat=1.18}
\usetikzlibrary{quotes,angles}
\usetikzlibrary{decorations.pathreplacing,positioning,shapes.geometric}
\usetikzlibrary{graphs,graphs.standard}
\usetikzlibrary{arrows,calc,patterns,intersections,fit,backgrounds,shadows}

\usetikzlibrary{external}
\tikzset{>=latex}
\tikzstyle{MLP}=[trapezium, trapezium angle=80, draw=black, thick, align=center, text=black, shape border rotate=-180, inner sep=3pt, outer sep=0pt]
\tikzstyle{block}=[draw=black, thick, align=center, rounded corners, minimum height=0.5cm, fill=red, text=black]

\pgfdeclarelayer{bg}    % declare background layer
\pgfsetlayers{bg,main}  % set the order of the layers (main is the standard layer)

\usetikzlibrary{fit}
\makeatletter
\tikzset{
  fitting node/.style={
    inner sep=0pt,
    fill=none,
    draw=none,
    reset transform,
    fit={(\pgf@pathminx,\pgf@pathminy) (\pgf@pathmaxx,\pgf@pathmaxy)}
  },
  reset transform/.code={\pgftransformreset}
}
\makeatother

\usepackage{siunitx}

\usepackage{csquotes}

\usepackage[capitalise,nameinlink]{cleveref}

\usepackage{glossaries-extra}

\setabbreviationstyle[acronym]{long-short}

\glssetcategoryattribute{acronym}{nohyperfirst}{true}

\newacronym{cir}{CIR}{Channel Impulse Response}
\newacronym{cnn}{CNN}{Convolutional Neural Network}
\newacronym{em}{EM}{electromagnetic}
\newacronym{gat}{GAT}{Geometric Attention Transformer}
\newacronym{gflownet}{GFlowNet}{Generative Flow Network}
\newacronym{gs}{GS}{Gaussian splatting}
\newacronym{gt}{GT}{ground truth}
\newacronym{mlp}{MLP}{multilayer perceptron}
\newacronym{nerf}{NeRF}{Neural Radiance Fields}
\newacronym{rf}{RF}{Radio Frequency}
\newacronym{ris}{RIS}{reconfigurable intelligent surface}
\newacronym{rx}{RX}{receiver}
\newacronym{sbr}{SBR}{shooting-and-bouncing-rays}
\newacronym{tx}{TX}{transmitter}
\newacronym{wrf}{WRF}{Wireless Radiation Field}

\usepackage{import}

\everymath=\expandafter{\the\everymath\displaystyle}

\author*[1]{\fnm{Jérome} \sur{Eertmans}}\email{jerome.eertmans@uclouvain.be}
\author[2]{\fnm{Enrico M.} \sur{Vitucci}}\email{enricomaria.vitucci@unibo.it}
\author[2]{\fnm{Vittorio} \sur{Degli-Esposti}}\email{v.degliesposti@unibo.it}
\author[3]{\fnm{Nicola} \sur{Di Cicco}}\email{nicola.dicicco@optit.net}
\author[1]{\fnm{Laurent} \sur{Jacques}}\email{laurent.jacques@uclouvain.be}
\author[1]{\fnm{Claude} \sur{Oestges}}\email{claude.oestges@uclouvain.be}

\affil*[1]{\orgdiv{Institute of Information and Communication Technologies, Electronics and Applied Mathematics}, \orgname{Université~catholique~de~Louvain}, \orgaddress{\city{Louvain-la-Neuve}, \country{Belgium}}}

\affil[2]{\orgdiv{Department of Electrical, Electronic, and Information Engineering}, \orgname{Università~di~Bologna}, \orgaddress{\city{Bologna}, \country{Italy}}}

\affil[3]{\orgname{OPTIT S.r.l.}, \orgaddress{\city{Bologna}, \country{Italy}}}

\begin{document}

%\title{Transform-Invariant Generative Ray Path Sampling for Efficient Radio Propagation Modeling with Point-to-Point Ray Tracing}

\title{Transform-Invariant Generative Ray Path Sampling for Efficient Radio Propagation Modeling}

%%=============================================================%%
%% GivenName  -> \fnm{Joergen W.}
%% Particle  -> \spfx{van der} -> surname prefix
%% FamilyName  -> \sur{Ploeg}
%% Suffix  -> \sfx{IV}
%% \author*[1,2]{\fnm{Joergen W.} \spfx{van der} \sur{Ploeg}
%%  \sfx{IV}}\email{iauthor@gmail.com}
%%=============================================================%%

\keywords{Channel Models, Generative Models, Machine Learning, Neural Networks, Radio Propagation, Ray Tracing}

%%\pacs[JEL Classification]{D8, H51}

%%\pacs[MSC Classification]{35A01, 65L10, 65L12, 65L20, 65L70}

\abstract{Ray tracing has become a standard for accurate radio propagation modeling, but suffers from exponential computational complexity, as the number of candidate paths scales with the number of objects raised to the interaction order. This bottleneck limits its use in large-scale or real-time applications, forcing traditional tools to rely on heuristics that reduce path candidates at the cost of potentially reduced accuracy. To overcome this limitation, we propose a machine-learning-assisted framework that replaces exhaustive path searching with intelligent sampling via Generative Flow Networks. Applying these generative models to this domain presents challenges, particularly sparse rewards due to the rarity of valid paths, which can lead to convergence failures and trivial solutions when evaluating high-order interactions in complex environments. To ensure robust learning and efficient exploration, our framework incorporates three key components. First, an \emph{experience replay buffer} captures and retains rare valid paths. Second, a uniform exploratory policy improves generalization and prevents overfitting to simple geometries. Third, a physics-based action masking strategy filters out physically impossible paths before the model considers them. Validated on idealized street-canyon scenarios, our model achieves substantial speedups over exhaustive search---up to $10\times$ faster on GPU and $100\times$ faster on CPU---while maintaining high coverage accuracy and successfully uncovering complex propagation paths. However, out-of-distribution evaluations on real-world Manhattan street geometries reveal that generalizing to substantially different urban morphologies requires further advancement in model capacity or alternative training strategies. Source code, tests, and a tutorial are available at \url{https://github.com/jeertmans/sampling-paths}.}

\maketitle

\section{Introduction}\label{sec:introduction}

Radio propagation modeling is at the foundation of modern telecommunications research and engineering, playing a crucial role in the design, optimization, and deployment of wireless communication systems~\cite{rappaport}. As wireless technologies continue to evolve, from macrocell deployments to highly dynamic local area networks, the demand for accurate, efficient, and scalable propagation models has become increasingly important~\cite{6g-wc-measurements-and-models}. The ability to predict how radio waves interact with complex environments directly impacts network coverage, capacity planning, interference mitigation, and overall system performance.

Traditional approaches to radio propagation modeling have long relied on empirical models and statistical methods that provide computational efficiency at the cost of physical accuracy~\cite{hata-model}. While these methods serve well for initial network planning and broad coverage estimation, they often fail to capture the intricate details of wave propagation in complex environments, particularly in dense urban areas, indoor spaces, or scenarios involving multiple reflections and diffractions~\cite{survey-models}. This limitation becomes increasingly problematic as modern applications demand precise channel characterization for advanced techniques such as massive MIMO, beamforming, and millimeter-wave communications.

Ray tracing is widely recognized as a highly effective method for accurate radio propagation modeling, offering a physics-based approach that can capture the complex interactions between electromagnetic waves and environmental objects with remarkable precision~\cite{rt-survey,rt-ref}. By modeling radio waves as rays that follow geometrical optics principles, ray tracing can account for multiple propagation mechanisms including reflection, refraction, diffraction, and scattering. This level of detail makes it invaluable for applications requiring high accuracy, such as site-specific channel modeling, interference analysis, and the design of advanced antenna systems, such as \glspl{ris}. Furthermore, these precise, physics-based simulations are essential for realizing the concept of a telecommunications \emph{digital twin}. By integrating ray tracing engines with high-resolution 3D environmental models, engineers can create high-fidelity virtual replicas of physical spaces, allowing for the dynamic simulation, testing, and optimization of wireless networks prior to real-world deployment.

However, the computational complexity of ray tracing remains its primary limitation~\cite{speed-up-rt}. The fundamental challenge lies in the exponential growth of potential ray paths as the number of reflections or diffractions increases and the environment becomes more complex. For point-to-point ray tracing---which aims to identify all possible propagation paths between a transmitter and a receiver---the number of path candidates to be evaluated grows exponentially with the maximum number of interactions considered. In practice, only a small fraction of these candidate paths actually contribute to the received signal, leading to substantial computational waste as the majority of processing time is spent evaluating invalid, i.e., physically obstructed, or insignificant paths. Point-to-point ray tracing (also referred to as \emph{exhaustive} ray tracing) thus becomes a computational bottleneck, particularly in large-scale or highly complex environments where exhaustive search is computationally intractable. As a result, methods like ray launching are used to limit the number of paths considered by launching a fixed number of rays and observing how they interact with the environment~\cite{rt-ref}. While these approaches are extremely computationally efficient for radio coverage map prediction, as the same rays can be reused to predict the signal received at multiple positions, they often require shooting a very large number of rays, typically much larger than the actual number of valid paths. Moreover, these techniques are sensitive to the angular spacing between rays, and introduce other caveats such as duplicate potential ray paths that should be discarded during post-processing. In this paper, however, our focus is on accelerating exhaustive point-to-point ray tracing, so exhaustive search remains the most relevant baseline.

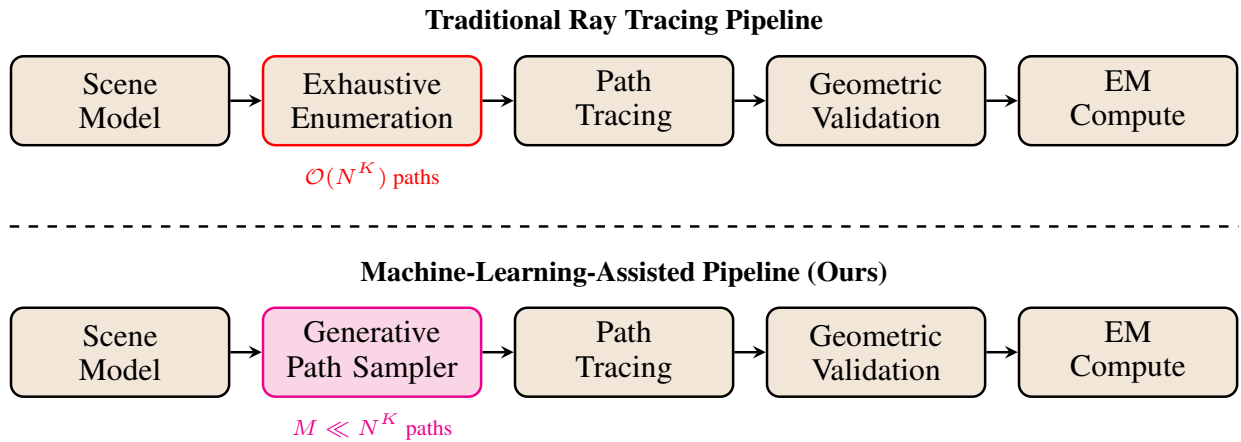
\begin{figure}[t]
  \centering
  % TODO: do not resize, make tikz figure fit page width directly
  \resizebox{\textwidth}{!}{
  \tikzsetnextfilename{tikzexternalize/pipeline}%
  \begin{tikzpicture}[
    node distance=1.0cm and 0.35cm,
    %font=\small,%\sffamily,
    >=stealth,
    % Styles
    block/.style={
      rectangle,
      draw=black,
      fill=brown!20,
      text width=2.2cm,
      align=center,
      minimum height=1cm,
      rounded corners,
      thick
    },
    ml_block/.style={
      block,
      draw=magenta,
      fill=magenta!20,
    },
    ex_block/.style={
      block,
      draw=red,
    },
    line/.style={draw, ->, thick},
    label_text/.style={font=\bfseries\footnotesize, align=left}
  ]

  % --- Traditional Pipeline (Top Row) ---
  \node[block] (t_scene) {Scene\\Model};
  \node[ex_block, right=of t_scene] (t_enum) {Exhaustive\\Enumeration};
  \node[block, right=of t_enum] (t_tra) {Path\\Tracing};
  \node[block, right=of t_tra] (t_val) {Geometric\\Validation};
  \node[block, right=of t_val] (t_em) {\glsxtrshort{em}\\Compute};

  % Arrows Top
  \draw[line] (t_scene) -- (t_enum);
  \draw[line] (t_enum) -- (t_tra);
  \draw[line] (t_tra) -- (t_val);
  \draw[line] (t_val) -- (t_em);

  % Label Top
  \node[above=0.1cm of t_tra, label_text] {Traditional Ray Tracing Pipeline};
  \node[below=0.05cm of t_enum, font=\scriptsize, color=red] {$\mathcal{O}(N^K)$ paths};

  % --- ML Pipeline (Bottom Row) ---
  % Position relative to top row to ensure alignment
  \node[block, below=1.8cm of t_scene] (m_scene) {Scene\\Model};
  \node[ml_block, right=of m_scene] (m_gen) {Generative\\Path Sampler};
  \node[block, right=of m_gen] (m_tra) {Path\\Tracing};
  \node[block, right=of m_tra] (m_val) {Geometric\\Validation};
  \node[block, right=of m_val] (m_em) {\glsxtrshort{em}\\Compute};

  % Arrows Bottom
  \draw[line] (m_scene) -- (m_gen);
  \draw[line] (m_gen) -- (m_tra);
  \draw[line] (m_tra) -- (m_val);
  \draw[line] (m_val) -- (m_em);

  % Label Bottom
  \node[above=0.1cm of m_tra, label_text] {Machine-Learning-Assisted Pipeline (Ours)};
  \node[below=0.05cm of m_gen, font=\scriptsize, color=magenta] {$M \ll N^K$ paths};

  % --- Visual Separator ---
  \draw[dashed, thick] ($(t_scene.south west)!0.5!(m_scene.north west) + (-0.0,0)$) -- ($(t_em.south east)!0.5!(m_em.north east) + (0.0,0)$);

\end{tikzpicture}%

  }
  \caption{Comparison of a traditional ray tracing pipeline versus our machine-learning-assisted approach. The core modification replaces the \textquote{Exhaustive Enumeration} bottleneck with an efficient \textquote{Generative Path Sampler,} drastically reducing the validation workload.}
  \label{fig:pipeline}
\end{figure}

Recent advances in machine learning have opened new avenues for addressing computational challenges across various domains, and radio propagation modeling is no exception~\cite{ml-radio-propa-survey1,ml-radio-propa-survey2}. The combination of increased computational power, sophisticated optimization algorithms, and automatic differentiation frameworks has enabled researchers to explore machine learning solutions for wireless channel prediction. The emergence of \gls{nerf} has particularly influenced this domain, with several recent works exploring \gls{nerf}-based approaches for wireless channel modeling~\cite{nerf2,r-nerf,nerf-apt,rf-rt-neural-objects}.

However, most existing approaches attempt to directly learn specific channel characteristics---such as path loss~\cite{wu2020artificial,thrane2020model}, received power~\cite{wi-gatr}, or coverage maps---rendering them highly dependent on specific environmental conditions, material properties, and frequency bands. Additionally, most of these approaches are limited to specific scenarios, such as indoor environments~\cite{wigsert} or specific frequency bands~\cite{thrane2020model}, and often require extensive retraining when applied to new environments or conditions.

This paper presents a fundamentally different approach: rather than learning the final channel characteristics, we propose enhancing the ray tracing process itself through machine-learning-assisted ray path sampling. Building upon our previous work~\cite{icmlcn2025}, we develop a comprehensive framework that leverages generative machine learning models to intelligently sample potential ray paths, significantly reducing the computational burden while maintaining the physical accuracy inherent in ray tracing methods.

Our approach addresses several key limitations of existing machine-learning-based propagation models. First, by working at the level of ray path generation rather than final channel prediction, our method preserves the fundamental physics of electromagnetic propagation and can be applied to compute various channel metrics from the same set of identified paths. Second, we ensure that our model exhibits proper invariance properties with respect to scene transformations (translation, rotation, and scaling), making it robust and generalizable across different environments. Third, our framework is designed to handle scenes of arbitrary complexity and size, learning general principles of ray propagation rather than memorizing specific environmental configurations.

Similar to~\cite{sandwich}, we treat ray path generation as a sequential decision-making problem, where a generative model learns to prioritize potentially valid paths among all possible candidates. This approach transforms the traditionally exhaustive search through an exponentially large space into an intelligent sampling process that focuses computational resources on the most promising path candidates. Moreover, trained via reinforcement learning, our model avoids the need for computationally expensive ground truth, and instead learns to prioritize the most promising paths based on their likelihood of contributing to the received signal. The result is a significant reduction in simulation time, both during training and during inference, while preserving the accuracy and physical interpretability of traditional ray tracing.

Despite these advancements, a critical question remains: can machine learning really benefit ray tracing users? With the recent rise in computational power and memory availability, traditional ray tracing methods have become increasingly viable for small to medium-sized scenes. For instance, modern differentiable ray tracing libraries like Sionna~RT~\cite{sionna-rt} can handle such scenarios efficiently. The theoretical turning points where standard ray tracing becomes computationally prohibitive due to memory constraints often remain distant for smaller scenes.
However, for large-scale environments, the exponential scaling of exhaustive ray tracing remains a significant limitation. While training such a machine learning model is complex and costly, requiring generalization across a class of scenes to be truly useful, the potential for massive speedups during inference is substantial. To be practical, the machine learning model must be lightweight, ensuring that its inference cost is negligible compared to the cost of exhaustive ray tracing.

Our major contributions are as follows.
\begin{enumerate}
  \item We introduce an improved machine-learning-assisted ray tracing framework that addresses the fundamental computational barrier of point-to-point ray tracing through intelligent path sampling.
  \item We develop a generative model architecture that exhibits proper invariance properties to scene transformations, ensuring robustness and generalizability across different environmental configurations, and linear inference complexity per sample with respect to scene size, representing a significant improvement in computational efficiency.
  \item We significantly enhance the robustness and training stability of our previous generative path sampling framework~\cite{icmlcn2025} through three key architectural refinements:
    \begin{enumerate}
      \item we resolve the \textbf{sparse-reward issue} by introducing a successful experience replay buffer;
      \item we replace the dropout mechanism with a \textbf{uniform exploratory policy} to suppress overfitting and improve generalization; and
      \item we implement a \textbf{physics-based action masking strategy} to drastically prune the search space.
    \end{enumerate}
    As quantified in \cref{sec:application}, these architectural improvements translate into concrete performance gains: specifically, they accelerate convergence, drastically increase the discovery rate of valid paths in complex geometries, and enable the effective learning of high-order interactions ($K \ge 2$) that were previously computationally prohibitive.
  \item We provide an open-source implementation using our own ray tracer DiffeRT~\cite{Eert2505:DiffeRT}, including full source code, test files, and a comprehensive tutorial\footnote{\url{https://differt.rtfd.io/npjwt2026/notebooks/sampling-paths.html}}, available at \url{https://github.com/jeertmans/sampling-paths}.
\end{enumerate}

The paper is organized as follows: \cref{sec:review} provides a comprehensive review of existing machine learning approaches to radio propagation modeling, highlighting their limitations and positioning our work within the broader literature. \Cref{sec:concept} introduces the fundamental concepts behind machine-learning-assisted ray tracing, outlining the computational challenges and our proposed solution. \Cref{sec:methodology} presents our methodology in detail, including the model architecture, training procedures, and theoretical analysis. \Cref{sec:application} demonstrates the practical application of our framework to radio coverage prediction in urban street canyon environments, providing comprehensive performance evaluation and comparison with traditional methods. Finally, we conclude with a discussion of implications, limitations, and future research directions.

\section{Related Work}\label{sec:review}

The use of machine learning in radio propagation modeling has accelerated in recent years, driven by advances in high-performance computing, the maturity and accessibility of automatic differentiation frameworks such as TensorFlow, PyTorch, and JAX~\cite{tensorflow2015-whitepaper,pytorch,jax2018github}, and the increasing availability of large-scale synthetic datasets~\cite{boston-twin,wigsert}.

The main motivation for applying machine learning in this domain is the classic trade-off between efficiency and accuracy. Traditional empirical models are fast but often too inaccurate for modern wireless systems~\cite{pred-models-survey}. On the other hand, full-wave \gls{em} solvers and high-fidelity ray tracing are precise but computationally expensive, scaling poorly with scene complexity. Machine learning offers a compelling compromise by approximating complex propagation phenomena at the speed of neural inference~\cite{winert,raypronet}. Moreover, the advent of differentiable programming opens the door to end-to-end optimization of wireless systems~\cite{sionna-rt}. Although native differentiable \gls{em} simulators like Sionna~RT~\cite{sionna-rt} and DiffeRT~\cite{Eert2505:DiffeRT} are emerging, most legacy tools are not differentiable. This has motivated research into machine-learning-based differentiable surrogates~\cite{winert,wi-gatr}.

This section reviews the existing literature, which we divide into two main categories: methods that learn channel characteristics directly, and those that serve as surrogates for specific components of the simulation pipeline. \Cref{tab:comparison} compares a few recent works on machine learning applied to radio propagation. We conclude by highlighting the key limitations that our work aims to address.

\begin{table}[htbp]
  \centering
  \caption{Comparison of recent works on machine learning applied to radio propagation, from earliest to latest.}
  \label{tab:comparison}
  \begin{tabular}{@{}llll@{}}
    \toprule
    Paper &
    Input(s) &
    Architecture &
    Output(s) \\ \midrule
    \cite{wu2020artificial} (2020) &
    \begin{tabular}[t]{@{}l@{}}Objects, \glsxtrshort{tx}, \glsxtrshort{rx},\\and \glsxtrshort{em} properties
    \end{tabular} &
    \glsxtrshort{mlp} &
    Path loss \\
    \cite{thrane2020model} (2020) &
    2D satellite images &
    \glsxtrshort{cnn} &
    Path loss \\
    \cite{nerf2} (2023) &
    \begin{tabular}[t]{@{}l@{}}\glsxtrshort{tx} and \glsxtrshort{rx}\\\textbf{(fixed scene)}
    \end{tabular} &
    \glsxtrshort{nerf} &
    Signal field \\
    \cite{winert} (2023) &
    Objects, \glsxtrshort{tx}, and \glsxtrshort{rx} &
    \glsxtrshort{mlp} &
    \begin{tabular}[t]{@{}l@{}}Multipath components\\(gain, delay, angles)
    \end{tabular} \\
    \cite{r-nerf} (2024) &
    \begin{tabular}[t]{@{}l@{}}Objects, \glsxtrshort{tx}, \glsxtrshort{rx}, and \glsxtrshort{ris}\\\textbf{(fixed scene)}
    \end{tabular} &
    \glsxtrshort{nerf} &
    Signal field re-radiated from \glsxtrshort{ris}\\
    \cite{sandwich} (2024) &
    \begin{tabular}[t]{@{}l@{}}Objects, \glsxtrshort{tx}, \glsxtrshort{rx},\\and \glsxtrshort{em} properties
    \end{tabular} &
    Transformer &
    \begin{tabular}[t]{@{}l@{}}Sequence of bounces\\(ray trajectories)
    \end{tabular} \\
    \cite{wi-gatr} (2024) &
    \begin{tabular}[t]{@{}l@{}}Objects, \glsxtrshort{tx}, \glsxtrshort{rx},\\and \glsxtrshort{em} properties
    \end{tabular} &
    \glsxtrshort{gat} &
    Received Power \\
    \cite{raypronet} (2024) &
    \begin{tabular}[t]{@{}l@{}}\glsxtrshort{tx} and \glsxtrshort{rx}\\\textbf{(fixed scene)}
    \end{tabular} &
    PointNet &
    Path loss \\
    \cite{nerf-apt} (2025) &
    \begin{tabular}[t]{@{}l@{}}\glsxtrshort{tx} and \glsxtrshort{rx}\\\textbf{(fixed scene)}
    \end{tabular} &
    \glsxtrshort{nerf} \& U-Net &
    Complex attenuation coefficients \\
    \cite{nerf-radio-map} (2025) &
    \begin{tabular}[t]{@{}l@{}}Objects, \glsxtrshort{tx}, \glsxtrshort{rx},\\ and \glsxtrshort{em} properties
    \end{tabular} &
    \glsxtrshort{mlp} &
    Complex attenuation coefficients \\
    \cite{wrf-gs,rf-3dgs} (2025--2026) &
    \begin{tabular}[t]{@{}l@{}}\glsxtrshort{tx} and \glsxtrshort{rx}\\\textbf{(fixed scene)}
    \end{tabular} &
    \glsxtrshort{nerf} \& 3D \glsxtrshort{gs} &
    Signal field \\
    \begin{tabular}[t]{@{}l@{}}\cite{icmlcn2025} (2025)\\and \textbf{this work}
    \end{tabular} &
    Objects, \glsxtrshort{tx}, and \glsxtrshort{rx} &
    \glsxtrshort{gflownet} &
    \begin{tabular}[t]{@{}l@{}}Path candidates \\(valid ray paths indices)
    \end{tabular} \\ \bottomrule
  \end{tabular}
\end{table}

\subsection{Direct Channel Characteristic Learning}

Much of the existing work on machine-learning-assisted propagation modeling frames the problem as a supervised learning task~\cite{ml-radio-propa-survey2}. In this approach, a neural network learns a mapping from environmental features directly to channel metrics like path loss or received power.

\textbf{Standard \glsxtrshort{mlp}-based Approaches:} Early work used standard \glspl{mlp} to predict channel characteristics from environmental features. For instance,~\cite{wu2020artificial} and~\cite{popoola2019determination} demonstrated reasonable accuracy in urban and very high frequency settings for path loss prediction. Similarly,~\cite{thrane2020model} presented a model-aided deep learning framework using a \gls{cnn} for path loss prediction at \qty{2.6}{\giga\hertz}. However, these approaches typically require extensive training data and are limited by their inability to generalize across different environments or operating conditions. They often struggle with the complex, non-linear relationships inherent in \gls{em} propagation and are inherently tied to the specific frequency and material properties of the training data.

\textbf{\glsxtrshort{nerf}-based Approaches:} Originally created for rendering novel views in computer graphics~\cite{nerf-rt}, \glspl{nerf} have recently been used for wireless channel modeling~\cite{nerf2,r-nerf,nerf-apt}. \Gls{nerf}\textsuperscript{2}~\cite{nerf2} uses a \textquote{turbo-learning} method that combines real and synthetic data to learn a continuous, volumetric representation of the scene for radio frequency propagation. R-\gls{nerf}~\cite{r-nerf} adapts this concept for \gls{ris}-assisted scenarios, employing a two-stage process to model field dynamics. More recently, \gls{nerf}-APT~\cite{nerf-apt} added attention mechanisms to an encoder-decoder architecture to better capture contextual information. Although these techniques can reconstruct radio environments with high fidelity, they are generally scene-specific and must be retrained from scratch for each new environment, a computationally expensive process. Recent 3D \gls{gs} variants, such as \gls{wrf}-\gls{gs} and \gls{rf}-3D\gls{gs}~\cite{wrf-gs,rf-3dgs}, push this idea further by achieving very fast training (few minutes) and query times (few milliseconds); however, they remain complementary to ray tracing rather than replacements for it, because they still rely on large amounts of scene data that are commonly generated or refined through ray tracing. In that sense, faster path sampling can directly support these downstream representations by reducing the cost of training-data generation.

\subsection{Alternative and Hybrid Approaches}

Given the challenges of end-to-end learning, some researchers have focused on improving specific parts of the propagation pipeline instead.

\textbf{Channel Impulse Response Estimation:} Rather than predicting a single metric like path loss, some models aim to predict the full \gls{cir}.  For example,~\cite{deep-rt-cir} developed a physics-informed network to estimate the \gls{cir} over a region, capturing both spatial and temporal channel characteristics.

\textbf{Neural Scene Representation:} Instead of replacing the simulator entirely, another strategy is to enrich the description of the environment. For example,~\cite{rf-rt-neural-objects} uses \glspl{nerf} to construct neural representations of complex objects. A standard ray tracing engine can then query these representations to account for scattering and diffraction effects that are difficult to model with traditional geometric primitives.

\textbf{Differentiable Simulation Surrogates:} The works most similar to ours are methods that view ray tracing as a decision-making process. The SANDWICH model~\cite{sandwich} is a differentiable, fully trainable surrogate for wireless ray tracing. It formulates the generation of ray trajectories as a sequential decision problem. However, SANDWICH aims to be a complete differentiable replacement for the simulator, whereas our approach focuses on optimizing \emph{path sampling} to speed up a conventional ray tracing engine.

\subsection{Limitations of Current Methods}

Despite these advances, current methods have fundamental limitations that prevent their widespread adoption in general-purpose radio planning tools.

\textbf{Complexity of Direct Field Learning:} Models that try to learn \gls{em} fields directly, whether using \glspl{mlp} or \glspl{nerf}, face a chaotic mapping problem. Tiny changes in geometry or frequency can lead to significant fluctuations in the electromagnetic field due to constructive and destructive interference.

Learning this high-frequency function demands large networks and vast amounts of data, which often results in overfitting.

\textbf{Poor Generalization and Lack of Invariance:} Most direct learning models are not invariant to simple geometric transformations like rotation or scaling. They also are tightly coupled to the specific frequencies and material properties used in their training data. A model trained for a \qty{2.4}{\giga\hertz} office environment is unlikely to work for a \qty{28}{\giga\hertz} street canyon without being completely retrained. To the best of our knowledge, only~\cite{wi-gatr} has demonstrated interest in geometrical invariance properties through the use of a \gls{gat}.

\textbf{Loss of Physical Interpretability:} End-to-end black-box models discard valuable ray-geometric information. For many 6G applications, like sensing, localization, and beam management, knowing \emph{which} paths (e.g., reflections, diffractions) contribute to the signal is just as important as the final signal strength.

\textbf{Computational Overhead:} While some machine-learning-based methods can be very accurate, inference often requires querying a large neural network. Paradoxically, this can be slower than highly optimized geometric ray tracing, especially in simple scenes. For a model to be practical, especially in real-time scenarios, its inference time must be negligible compared to that of the physics engine it assists~\cite{realtime-neural-receiver}.

\textbf{The Need for Intelligent Sampling:} The limitations discussed above suggest that the role of machine learning should be to guide the physics engine, not replace it. If we can learn to \emph{identify} valid ray paths, rather than predict their impact on the received signal, we can decouple the learning problem from frequency and material properties. This would allow a model to learn the general, geometric principles of propagation while leaving the precise field calculations to a robust, physics-based solver.

\section{Principle of Machine-Learning-Assisted Ray Tracing}\label{sec:concept}

As established in previous sections, the primary limitation of applying exhaustive point-to-point ray tracing to complex scenarios is the combinatorial explosion of potential paths. This section presents the fundamental concept of our solution: a machine-learning-assisted framework that transforms path discovery from a brute-force search into an intelligent sampling process. Rather than attempting to learn the final \gls{em} fields directly---which often leads to the generalization issues noted in~\cref{sec:review}---we use machine learning to guide the ray tracing engine itself. This approach preserves the physical accuracy of the simulation while drastically improving its efficiency.

\subsection{From Exhaustive Search to Intelligent Sampling}

\begin{figure}[htbp]
  \centering
  \tikzsetnextfilename{tikzexternalize/complete_graph}%
  \begin{tikzpicture}
  \graph[simple, edges={<->,opacity=0.5}, nodes={draw=none, circle, minimum width=5mm}, empty nodes, phase=180] { subgraph K_n [V={1,2,3,4,5,6},clockwise,radius=2cm];
    1 --[<->,dashed] 5;
    2 --[<->,dashed] 5;
    3 --[<->,dashed] 5;
    4 --[<->,dashed] 5;
    6 --[<->,dashed] 5;
    1 --[->, red, thick, opacity=1] 6;
  };
  \path (1) -- (4) node[midway] (center) {};
  \path (1) -- ++(-2,0) node[draw, circle, inner sep=1pt] (tx) {\glsxtrshort{tx}};
  \path (4) -- ++(+2,0) node[draw, circle, inner sep=1pt] (rx) {\glsxtrshort{rx}};
  \foreach \x [count=\idx from 0] in {1,...,6} {
    \pgfmathparse{180 - \idx * (360 / 6)}
    \ifnum\x=5
    \node[opacity=0.5] at ($(center)+(\pgfmathresult:2cm)$) {...};
    \else
    \ifnum\x=1
    \node[draw, circle, inner sep=0pt, text width=5mm, align=center, red] at ($(center)+(\pgfmathresult:2cm)$) {$o_\x$};
    \else
    \ifnum\x=6
    \node[draw, circle, inner sep=0pt, text width=5mm, align=center, red] at ($(center)+(\pgfmathresult:2cm)$) {$o_N$};
    \else
    \node[draw, circle, opacity=0.5, inner sep=0pt, text width=5mm, align=center] at ($(center)+(\pgfmathresult:2cm)$) {$o_\x$};
    \fi
    \fi
    \fi

    \ifnum\x=4
    \else
    \ifnum\x=5
    \draw[->,dashed, opacity=0.5] (tx) -- (\x);
    \else
    \draw[->,opacity=0.5] (tx) -- (\x);
    \fi
    \fi

    \ifnum\x=1
    \else
    \ifnum\x=5
    \draw[<-,dashed, opacity=0.5] (rx) -- (\x);
    \else
    \draw[<-,opacity=0.5] (rx) -- (\x);
    \fi
    \fi
  };
  \draw[->,opacity=0.5](tx) to[bend left] (4);
  \draw[<-,opacity=0.5] (rx) to[bend left] (1);
  \draw[->,opacity=0.5] (tx) to[bend left] (rx);
  \draw[->, red, thick] (tx) -- (1);
  \draw[->, red, thick] (6) -- (rx);
\end{tikzpicture}%

  \caption{Representation of all possible path candidates from \gls{tx} to \gls{rx} in a scene with $N$ objects. Each path corresponds to a sequence of object interactions, forming a directed graph where exhaustive search explores all branches. An example path candidate, highlighted in red, is $\mathrm{\glsxtrshort{tx}}\to o_1 \to o_N \to \mathrm{\glsxtrshort{rx}}$.}
  \label{fig:complete_graph}
\end{figure}
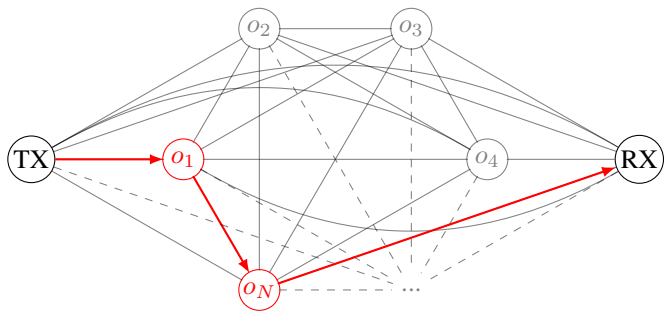

A traditional point-to-point ray tracing algorithm must test every possible sequence of object interactions up to a given order $K$. For a scene with $N$ objects, denoted $o_1$ to $o_N$, this results in up to $\mathcal{O}(N^K)$ candidate paths, the vast majority of which are physically invalid as a result of being blocked by other objects. Searching for all possible path candidates amounts to testing all the path candidates from \gls{tx} to \gls{rx} in the directed graph illustrated on \cref{fig:complete_graph}. This exponential scaling represents a fundamental bottleneck.

Our core insight is to reframe this exhaustive search as a \emph{sequential decision-making problem}. Instead of viewing a path as a geometrical entity to be tested, we treat it as a sequence of choices. Starting from \gls{tx}, we select the first object for interaction, then the next, continuing until we reach the desired number of interactions, after which we attempt to connect to \gls{rx}. This process forms a decision tree where each branch represents a potential path. The goal is to learn a strategy that intelligently explores this tree, prioritizing branches likely to yield valid paths while pruning the rest.

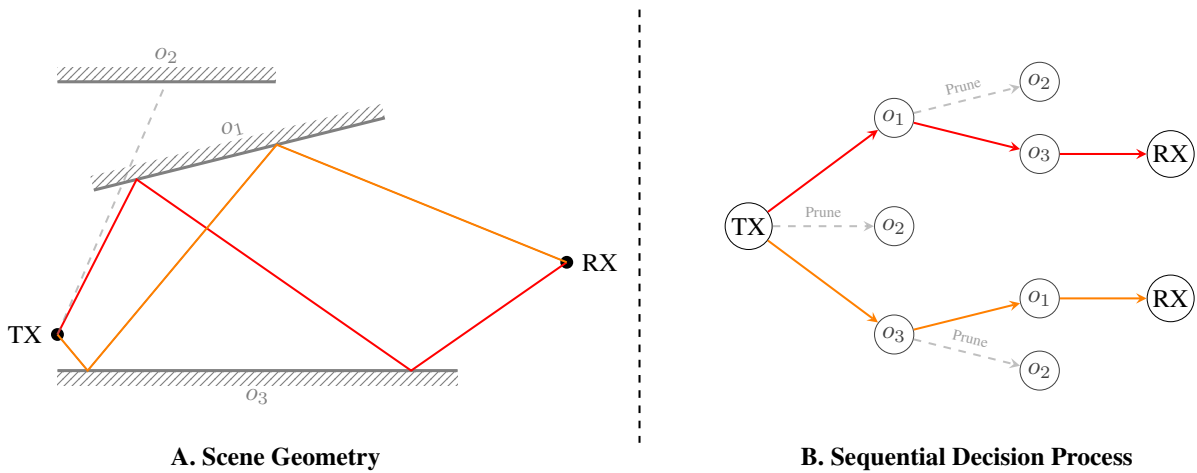
\begin{figure}
  \centering
  % TODO: do not resize, make tikz figure fit page width directly
  \resizebox{\textwidth}{!}{
  \tikzsetnextfilename{tikzexternalize/sequential_decision}%
  \begin{tikzpicture}[
    %font=\footnotesize,%\sffamily,
    >=stealth,
    % --- Styles ---
    % TX/RX
    tx/.style={circle, draw=black, thick, fill=black, inner sep=1.5pt, label={left:\glsxtrshort{tx}}},
    rx/.style={circle, draw=black, thick, fill=black, inner sep=1.5pt, label={right:\glsxtrshort{rx}}},
    % Mirrors
    mirror line/.style={draw, very thick, black!50},
    mirror hatch/.style={pattern=north east lines, pattern color=black!50},
    % Paths
    valid_path_1/.style={-, red, thick}, % Style for Path 1
    valid_path_2/.style={-, orange, thick},  % Style for Path 2
    invalid_path/.style={-, gray, thick, dashed, opacity=0.5},
    % Tree Styles
    tree_node/.style={draw, circle, opacity=0.75, inner sep=0pt, text width=5mm, align=center},
    tree_root/.style={tree_node, inner sep=1pt,opacity=1},
    tree_leaf/.style={tree_root},
    edge_valid/.style={->, red, thick},
    edge_valid_2/.style={->, orange, thick},
    edge_invalid/.style={->, gray, thick, dashed, opacity=0.5},
    prune_label/.style={midway, above, sloped, font=\tiny, text=gray, opacity=0.75}
  ]

  % ==============================================================================
  % LEFT SCOPE: Realistic Scene Geometry with Specular Reflections
  % ==============================================================================
  \begin{scope}[local bounding box=scene_box]
    % --- Define Coordinates ---
    \coordinate (TX_c) at (1, 1.0);
    \coordinate (RX_c) at (8, 2.0);

    % Surface O1 (Top Right slanted)
    \coordinate (O1_A) at (1.5, 3.0); \coordinate (O1_B) at (5.5, 4.0);
    % Surface O2 (Top Left flat - Invalid)
    \coordinate (O2_A) at (1, 4.5); \coordinate (O2_B) at (4, 4.5);
    % Surface O3 (Bottom flat)
    \coordinate (O3_A) at (1.0, 0.5); \coordinate (O3_B) at (6.5, 0.5);

    % --- Draw Mirrors ---
    % Define paths for intersections
    \path[name path=surf_o_1] (O1_A) -- (O1_B);
    \path[name path=surf_o_2] (O2_A) -- (O2_B);
    \path[name path=surf_o_3] (O3_A) -- (O3_B);

    % Visual representation
    \foreach \a/\b/\name/\pos/\hatchangle in {
      O1_A/O1_B/o_1/above/90,
      O2_A/O2_B/o_2/above/90,
      O3_A/O3_B/o_3/below/-90
    } {
      \fill[mirror hatch] (\a) -- (\b) -- ($(\b)!-0.2cm!\hatchangle:(\a)$) -- ($(\a)!-0.2cm!-\hatchangle:(\b)$) -- cycle;
      \draw[mirror line] (\a) -- (\b) node[midway, \pos=1.2mm, sloped, text=black!50] {$\name$};
    }

    \node[tx] (TX) at (TX_c) {};
    \node[rx] (RX) at (RX_c) {};

    % --- GEOMETRIC PATH CALCULATIONS ---

    % --- Path 1: TX -> o1 -> o3 -> RX ---
    % 1. Image of TX across o1
    \tkzDefPointBy[reflection=over O1_A--O1_B](TX) \tkzGetPoint{TX_i1};
    % 2. Image of TX_i1 across o3
    \tkzDefPointBy[reflection=over O3_A--O3_B](TX_i1) \tkzGetPoint{TX_i1_i3};
    % 3. Intersection on o3 (P1_2)
    \path[name path=virt_p1_b,overlay] (TX_i1_i3) -- (RX_c);
    \path[name intersections={of=surf_o_3 and virt_p1_b, by=P1_2}];
    % 4. Intersection on o1 (P1_1)
    \path[name path=virt_p1_a] (TX_i1) -- (P1_2);
    \path [name intersections={of=surf_o_1 and virt_p1_a, by=P1_1}];
    % 5. Draw Path 1
    \draw[valid_path_1] (TX_c) -- (P1_1) -- (P1_2) coordinate[midway](mid1) -- (RX_c);

    % --- Path 2: TX -> o3 -> o1 -> RX ---
    % 1. Image of TX across o3
    \tkzDefPointBy[reflection=over O3_A--O3_B](TX) \tkzGetPoint{TX_i3};
    % 2. Image of TX_i3 across o1 (Changed target to o1)
    \tkzDefPointBy[reflection=over O1_A--O1_B](TX_i3) \tkzGetPoint{TX_i3_i1};
    % 3. Intersection on o1 (P2_2)
    \path[name path=virt_p2_b] (TX_i3_i1) -- (RX_c);
    \path [name intersections={of=surf_o_1 and virt_p2_b, by=P2_2}];
    % 4. Intersection on o3 (P2_1)
    \path[name path=virt_p2_a] (TX_i3) -- (P2_2);
    \path[name intersections={of=surf_o_3 and virt_p2_a, by=P2_1}];
    % 5. Draw Path 2
    \draw[valid_path_2] (TX_c) -- (P2_1) -- (P2_2) coordinate[midway](mid2) -- (RX_c);

    % --- Invalid Path Hint (TX -> o2) ---
    \draw[invalid_path] (TX_c) -- ($(O2_A)!0.5!(O2_B)$);

    \node[anchor=south, font=\bfseries] at (4, -1.0) {A. Scene Geometry};
  \end{scope}

  % Vertical Separator
  \draw[dashed, thick] (9.0, 5.5) -- (9.0, -0.5);

  % ==============================================================================
  % RIGHT SCOPE: Decision Tree
  % ==============================================================================
  \begin{scope}[xshift=10.5cm, yshift=2.5cm, local bounding box=tree_box]

    % Level 0
    \node[tree_root] (T0) at (0,0) {\glsxtrshort{tx}};

    % Level 1
    \node[tree_node] (T1) at (2.0, 1.5) {$o_1$}; % Leads to Path 1
    \node[tree_node] (T2) at (2.0, 0) {$o_2$};   % Invalid (Prune)
    \node[tree_node] (T3) at (2.0, -1.5) {$o_3$}; % Leads to Path 2

    % Edges L0 -> L1
    \draw[edge_valid] (T0) -- (T1);
    \draw[edge_invalid] (T0) -- (T2) node[prune_label] {Prune};
    \draw[edge_valid_2] (T0) -- (T3);

    % --- Branch 1: TX -> o1 -> o3 ---
    \node[tree_node] (T11) at (4.0, 2.0) {$o_2$}; % Prune
    \node[tree_node] (T12) at (4.0, 1.0) {$o_3$}; % Valid
    \draw[edge_invalid] (T1) -- (T11) node[prune_label] {Prune};
    \draw[edge_valid] (T1) -- (T12);
    % Leaf
    \node[tree_leaf] (T12R) at (5.8, 1.0) {\glsxtrshort{rx}};
    \draw[edge_valid] (T12) -- (T12R);

    % --- Branch 2: TX -> o3 -> o1 ---
    \node[tree_node] (T31) at (4.0, -1.0) {$o_1$}; % Valid (Changed from o2 to o1)
    \node[tree_node] (T32) at (4.0, -2.0) {$o_2$}; % Prune
    \draw[edge_valid_2] (T3) -- (T31);
    \draw[edge_invalid] (T3) -- (T32) node[prune_label] {Prune};
    % Leaf
    \node[tree_leaf] (T31R) at (5.8, -1.0) {\glsxtrshort{rx}};
    \draw[edge_valid_2] (T31) -- (T31R);

    \node[anchor=south, font=\bfseries] at (3, -3.5) {B. Sequential Decision Process};
  \end{scope}

\end{tikzpicture}%

  }
  \caption{Illustration of pathfinding as a sequential decision process. A machine learning model learns to navigate the decision tree (right), sampling interaction sequences (e.g., $\mathrm{\glsxtrshort{tx}}\to o_3 \to o_1 \to \mathrm{\glsxtrshort{rx}}$) that correspond to valid geometric paths in the scene (left) while avoiding the expensive exploration of invalid branches.}
  \label{fig:sequential_decision}
\end{figure}

\Cref{fig:sequential_decision} illustrates this concept. Instead of exhaustively evaluating all combinations (e.g., $\mathrm{\glsxtrshort{tx}}\to o_1 \to o_2 \to \mathrm{\glsxtrshort{rx}}$, $\mathrm{\glsxtrshort{tx}}\to o_1 \to o_3 \to \mathrm{\glsxtrshort{rx}}$, ...), the model learns from the scene geometry that paths via $o_1$ and $o_3$ are promising, whereas paths involving $o_2$ are invalid and can be disregarded.

While pruning the decision tree is a well-established concept in ray tracing (often using visibility-based heuristics)~\cite{visibility-1,visibility-2}, determining exact visibility is computationally expensive, often more so than checking the paths themselves. Consequently, approximate methods are typically used. We propose learning a probabilistic model to rapidly estimate which branches of the decision tree warrant exploration based on global scene geometry. This is combined with fast, approximate visibility checks to ensure hard geometric constraints must always be satisfied.

The objective is to train a generative model to act as a highly efficient path sampler. This model learns a probability distribution $P(\mathbf{p}|\mathbf{s})$ over path candidates $\mathbf{p}$ given a scene description $\mathbf{s}$ (i.e., a set of points including \gls{tx} and \gls{rx} position vectors, plus the triangle facets). A properly trained model assigns high probabilities to physically valid paths, allowing us to recover them by sampling a small number of candidates, $M$, where $M \ll N^K$. The sampling budget, $M$, can be adjusted to trade off computational cost and accuracy, enabling significant speedups while maintaining high-fidelity ray tracing results.

\subsection{Key Advantages of the Machine-Learning-Assisted Approach}

The shift from exhaustive enumeration to guided path sampling offers several critical advantages that address the limitations of both traditional ray tracing and existing \textquote{black box} machine learning methods.

\subsubsection{Generalization and Invariance}

A key design goal is to develop a model that generalizes across a \emph{class} of scenes (e.g., urban street canyons) rather than overfitting to a single environment. We achieve this by learning fundamental geometric propagation principles related to inter-object visibility, which remain constant, rather than frequency- or material-dependent field values. This is enforced through:
\begin{itemize}
  \item \textbf{Transform Invariance:} The model is invariant to scene translations, azimuthal rotations, and scaling, ensuring robustness to changes in the coordinate system.
  \item \textbf{Permutation Invariance:} The model treats scene objects as an unordered set, making the output independent of the arbitrary ordering of objects in the input data. This allows the framework to handle scenes of arbitrary size and complexity.
\end{itemize}

Those points are detailed in \cref{sec:methodology}.

\subsubsection{Preservation of Physics and Interpretability}

Since our model only replaces the path candidate generation step, the final validation and electromagnetic computation are performed by the standard, physically grounded ray tracing engine. We therefore retain the full accuracy and interpretability of traditional ray tracing. Crucially, we can still obtain detailed geometric path information (e.g., angles of arrival/departure, delay), which is essential for many applications but typically lost in end-to-end learning models.

\subsubsection{Computational Cost Reduction}

The primary benefit is a reduction in computational complexity. When using an exhaustive search, the runtime is approximately $\mathcal{O}\left(N^K \cdot C_{\text{validation}}\right)$, where $C_{\text{validation}}$ is the validation cost per path. Our method's cost is $\mathcal{O}\left(M \cdot\left(C_{\text{inf}} +  C_{\text{validation}}\right)\right)$, where $C_{\text{inf}}$ is the inference cost per path and $M$ is the number of sampled paths.

The practical speedup depends on the hardware architecture:
\begin{itemize}
  \item \textbf{CPU (Serial):} Since operations are largely serial, the speedup is significant provided $M \ll N^K$ and the model is lightweight ($C_{\text{inf}}$ is negligible).
  \item \textbf{GPU (Parallel):} Modern hardware, like GPUs, validates paths in large batches ($B$), where validating one million ray paths takes the same amount of time as validating one path. For small scenes where $N^K \le B$, exhaustive ray tracing is already fast. Moreover, approximate visibility heuristics can reduce the number of candidates, making almost-exhaustive\footnote{Some heuristics may occasionally prune valid path candidates.} ray tracing affordable on medium-sized scenes, such as small cities. However, for large-scale scenes where $N^K > B$, exhaustive search becomes memory-bound and requires multiple slow batches. By sampling the most important paths within a single batch ($M \le B$), our model bypasses this limitation, offering massive scalability for complex environments.
\end{itemize}

In \cref{sec:benchmarks}, we provide a detailed cost analysis of our method compared to traditional ray tracing and highlight the scenarios where our approach offers the most significant advantages, and where it may be less beneficial.

\subsection{Integration into the Ray Tracing Pipeline}
Our method serves as a drop-in replacement for the path generation stage of a conventional ray tracing pipeline. This modularity ensures seamless integration with existing, optimized ray tracing tools. The modified workflow is depicted in \cref{fig:pipeline}. All stages beyond path generation remain identical, guaranteeing that the final output retains high physical fidelity.

\section{Methodology}
\label{sec:methodology}

In this section, we present the proposed generative framework for sampling valid radio propagation paths. Building upon our preliminary work~\cite{icmlcn2025}, we model the path tracing problem as a discrete sequential decision process solved via a \gls{gflownet}~\cite{bengio2023gflownet}. We introduce a rigorous geometric pre-processing pipeline to ensure spatial invariance and detail the training objective, which incorporates flow matching loss and improved training strategies such as uniform exploratory policy and successful experience replay buffer.

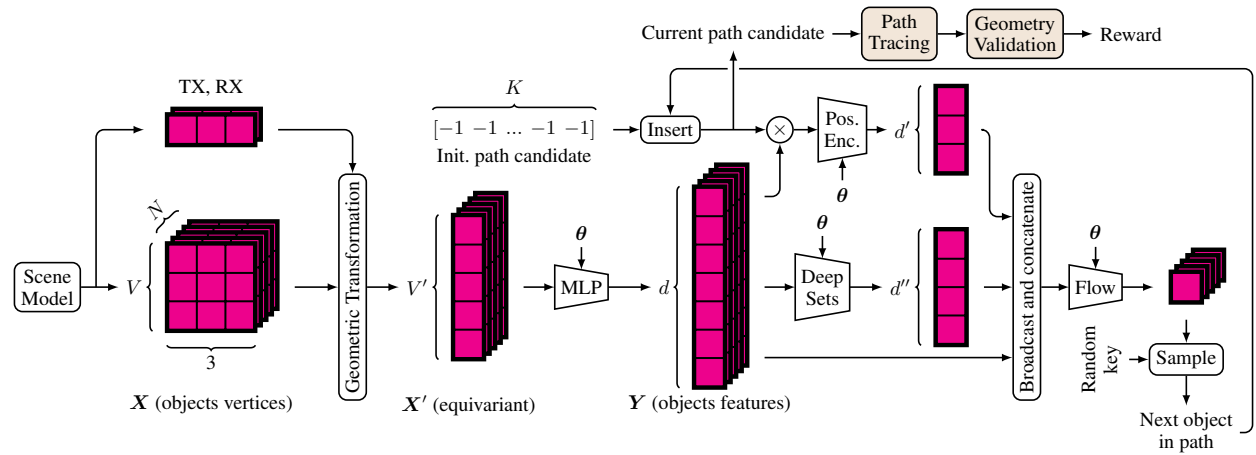
\begin{figure}[htbp]
  \centering
  % TODO: do not resize, make tikz figure fit page width directly
  \resizebox{\textwidth}{!}{
  \tikzsetnextfilename{tikzexternalize/model}%
  \begin{tikzpicture}
  \newcommand\Nobjects{5}
  \pgfkeys{
    /tensor/.is family, /tensor,
    default/.style =
    {
      m = 1,
      n = 1,
      d = 1,
      x scale = 1,
      y scale = 1,
      x step = 0.5,
      y step = 0.5,
      dx = +0.02,
      dy = +0.02,
      dz = -0.20,
      fill = magenta,
      line color = black,
      line width = 1pt,
      border color = black,
      border width = 2pt,
      name = none,
      anchor = center,
      at = center,
    },
    m/.estore in = \Tm,
    n/.estore in = \Tn,
    d/.estore in = \Td,
    x scale/.estore in = \TxScale,
    y scale/.estore in = \TyScale,
    x step/.estore in = \TxStep,
    y step/.estore in = \TyStep,
    dx/.estore in = \Tdx,
    dy/.estore in = \Tdy,
    dz/.estore in = \Tdz,
    fill/.estore in = \Tfill,
    line color/.estore in = \TlineColor,
    line width/.estore in = \TlineWidth,
    border color/.estore in = \TborderColor,
    border width/.estore in = \TborderWidth,
    name/.estore in = \Tname,
    anchor/.estore in = \Tanchor,
    at/.estore in = \Tat,
  }

  \newcommand\tensor[1][]{%
    \pgfkeys{/tensor,default,#1}
    \ifnum\Td>0
    \foreach \k in {\Td,...,1} {
      \pgfmathtruncatemacro\i{\k-1}
      \node[rectangle,anchor=\Tanchor,fill=\Tfill,inner sep=0,shift={(\i*\Tdx,\i*\Tdy,\i*\Tdz)}] (\Tname\i) at (\Tat) {\tikz{\draw[line width=\TlineWidth,color=\TlineColor,fill=\Tfill] (0,0) grid[xstep=\TxStep,ystep=\TyStep] +({(\Tn *
      \TxScale * \TxStep)},{(\Tm * \TyScale * \TyStep)});}};
      \draw[line width=\TborderWidth, color=\TborderColor] (\Tname\i.south
      west) rectangle (\Tname\i.north east);
    }
    \fi
  }

  \node[rectangle,draw,rounded corners,thick,align=center] (input) {Scene\\Model};

  \coordinate (ref) at ($(input.east) + (1.5,0)$);
  \tensor[m=3,n=3,name=X,d=\Nobjects,anchor=west,at=ref];
  \coordinate (ref) at ($(X0.north west) + (0,2)$);
  \tensor[m=1,n=3,name=tx_rx,d=2,anchor=west,at=ref];

  \node[above] at ([yshift=2mm]tx_rx0.north) {\glsxtrshort{tx}, \glsxtrshort{rx}};

  \draw [
    thick,
    decoration={
      brace,
      mirror,
      raise=0.25cm
    },
    decorate
  ] (X0.south west) -- (X0.south east)
  node [pos=0.5,anchor=north,yshift=-0.30cm] {3};

  \draw [
    thick,
    decoration={
      brace,
      mirror,
      raise=0.25cm
    },
    decorate
  ] (X0.north west) -- (X0.south west)
  node [pos=0.5,anchor=east,xshift=-0.30cm] (v) {$V$};

  \draw[->,thick] (input) -- (v) node[pos=.4] (mid) {};
  \draw[->,thick,rounded corners] (mid.center) |- ([xshift=-4mm]tx_rx0.west);

  \draw [
    thick,
    decoration={
      brace,
      mirror,
      raise=0.25cm
    },
    decorate
  ] (X4.north west) -- (X0.north west)
  node [pos=0.5,sloped,anchor=south,yshift=+0.3cm] {$N$};

  \path (X0.east) -- ++(1.7,0) node[rectangle,draw,rounded corners,thick,align=center,rotate=90] (geom_x) {Geometric Transformation};
  \draw[<-,thick] (geom_x) -- ([xshift=3mm]X4.east |- X0.east);
  \draw[<-,thick,rounded corners] (geom_x.east) |- ([xshift=3mm]tx_rx1.east |- tx_rx0.east);

  \coordinate (ref) at ($(geom_x.south) + (1.5,0)$);
  \tensor[m=5,n=1,name=XP,d=\Nobjects,anchor=west,at=ref];

  \draw [
    thick,
    decoration={
      brace,
      mirror,
      raise=0.25cm
    },
    decorate
  ] (XP0.north west) -- (XP0.south west)
  node [pos=0.5,anchor=east,xshift=-0.30cm] (vp) {$V'$};

  \draw[->,thick] (geom_x) -- (vp);

  \path (XP0.east) -- ++(1.7,0) node[MLP] (mlp_xp) {MLP};
  \coordinate (ref) at ($(mlp_xp.east) + (1.5,0)$);
  \tensor[m=7,n=1,name=Y,d=\Nobjects,anchor=west,at=ref]
  \draw[<-,thick] (mlp_xp) -- ([xshift=3mm]XP4.east |- XP0.east);
  \draw[<-,thick] (mlp_xp.north) -- ++(0,.4) node[above] {$\boldsymbol{\theta}$};

  \draw [
    thick,
    decoration={
      brace,
      mirror,
      raise=0.25cm
    },
    decorate
  ] (Y0.north west) -- (Y0.south west)
  node [pos=0.5,anchor=east,xshift=-0.30cm] (d) {$d$};

  \draw[->,thick] (mlp_xp) -- ([xshift=-3mm]d);

  \node[below] at (X0.south |- Y0.south) {$\boldsymbol{X}$ (objects vertices)};
  \node[below] at (XP0.south |- Y0.south) {$\boldsymbol{X}'$ (equivariant)};
  \node[below] at (Y0.south) {$\boldsymbol{Y}$ (objects features)};

  \path (Y0.east) -- ++(1.7,0) node[MLP] (deep_sets) {Deep\\Sets};
  \draw[<-,thick] (deep_sets.north) -- ++(0,.4) node[above] {$\boldsymbol{\theta}$};
  \draw[<-,thick] (deep_sets) -- ([xshift=3mm]Y4.east |- Y0.east) node[pos=1] (end) {} node[midway] (mid) {};
  \path (Y0.south) -- (Y0.north) node[pos=0.15] (Y025) {} node[pos=0.95] (Y075) {};

  \coordinate (ref) at ($(deep_sets.east) + (1.5,0)$);
  \tensor[m=4,n=1,name=scene_feats,d=1,anchor=west,at=ref]

  \draw [
    thick,
    decoration={
      brace,
      mirror,
      raise=0.25cm
    },
    decorate
  ] (scene_feats0.north west) -- (scene_feats0.south west)
  node [pos=0.5,anchor=east,xshift=-0.30cm] (dpp) {$d''$};
  \draw[->,thick] (deep_sets) -- (dpp);

  % TODO: undo xshift hack and place insert first
  \node[inner sep=0pt,label=below:{Init. path candidate}] (pc_init) at ([xshift=8mm]XP0.center |- tx_rx0.center) {$
    \begin{bmatrix}-1 & -1 & ... & -1 & -1
  \end{bmatrix}$};

  \draw [
    thick,
    decoration={
      brace,
      mirror,
      raise=0.15cm
    },
    decorate
  ] (pc_init.north east) -- (pc_init.north west)
  node [pos=0.5,anchor=south,yshift=+0.30cm] {$K$};

  \path (pc_init.east) -- ++(1.3,0) node[draw,rounded corners,thick] (insert) {Insert};
  \draw[<-,thick] (insert) -- ([xshift=3mm]pc_init.east);

  \path[->,thick,rounded corners] (end.center |- Y075.center) -| (mid.center |- insert.center) node[circle, draw=black, thick, inner sep=1pt, font=\bfseries] (mul_feats) {$\times$};
  \draw[->,thick,rounded corners] (end.center |- Y075.center) -| (mul_feats);
  \draw[->,thick] (insert) -- (mul_feats) node[pos=0.5] (mid) {};
  \draw[->,thick] (mid.center) -- ++(0,1.4) node[above] (pc) {Current path candidate};
  \draw[->,thick] (pc.east) -- ++(.5cm,0) node[rounded corners,draw,thick,align=center,right,fill=brown!20] (pt) {Path\\Tracing};
  \draw[->,thick] (pt.east) -- ++(.5cm,0) node[rounded corners,draw,thick,align=center,right,fill=brown!20] (gv) {Geometry\\Validation};
  \draw[->,thick] (gv.east) -- ++(.5cm,0) node[right] (r) {Reward};

  \path (mul_feats.east) -- ++(0.85,0) node[MLP] (pos_enc) {Pos.\\Enc.};
  \draw[<-,thick] (pos_enc.south) -- ++(0,-.4) node[below] {$\boldsymbol{\theta}$};
  \draw[->,thick] (mul_feats) -- (pos_enc);

  \coordinate (ref) at (scene_feats0.west |- pos_enc);
  \tensor[m=3,n=1,name=pc_feats,d=1,anchor=west,at=ref]

  \draw [
    thick,
    decoration={
      brace,
      mirror,
      raise=0.25cm
    },
    decorate
  ] (pc_feats0.north west) -- (pc_feats0.south west)
  node [pos=0.5,anchor=east,xshift=-0.30cm] (dp) {$d'$};
  \draw[->,thick] (pos_enc) -- (dp);

  %\draw[->] (($(deep_sets.east) + (0.75,.3)$) -- ++(1.2,0) node[circle, draw=black, thick, inner sep=1pt, font=\bfseries] (add_feats) {+};

  \path (scene_feats0.east) -- ++(1.05,0) node[rectangle,draw,rounded corners,thick,align=center,rotate=90] (concat) {Broadcast and concatenate};

  \path (concat.south) -- ++(0.95,0) node[MLP] (flow) {Flow};
  \draw[<-,thick] (flow.north) -- ++(0,.4) node[above] {$\boldsymbol{\theta}$};
  \draw[->,thick] (concat) -- (flow);
  \draw[<-,thick] (concat) -- ([xshift=3mm]scene_feats0.east);

  \path (scene_feats0.east) -- (concat.north) node[pos=0.5] (midway_bend) {};
  \path (concat.west) -- (concat.east) node[pos=0.25] (concat25) {} node[pos=0.75] (concat75) {};

  \draw[->,thick,rounded corners] (end.center |- Y025.center) -- (concat.north |- Y025.center) node[pos=1] (hit) {};

  \path (concat.north) -- ++($(concat.north) - (hit.center)$) node[pos=1,] (target) {};

  \draw[->,thick,rounded corners] ([xshift=3mm]pc_feats0.east) -- (midway_bend.center |- pc_feats0.east) |- (target.center);

  \coordinate (ref) at ($(flow.east) + (0.85,0)$);
  \tensor[m=1,n=1,name=flows,d=\Nobjects,anchor=west,at=ref]
  \draw[->,thick] (flow) -- ([xshift=-3mm]flows0.west);

  \node[draw,rounded corners,thick,align=center] (sample) at ($(flows0.south) + (0,-1)$) {Sample};
  \node[align=center] (object_idx) at ([yshift=-10mm]sample.south) {Next object\\in path};

  \draw[->,thick] ([yshift=-3mm]flows0.south) -- (sample);
  \draw[->,thick] (sample) -- (object_idx);
  \draw[rounded corners,thick] (object_idx) -| ([xshift=5mm,yshift=+3mm]sample.east |- pc_feats0.north) coordinate (tmp) -- ([xshift=1mm]pc.center |- tmp);
  \draw[->,thick,rounded corners] ([xshift=-1mm]pc.center |- tmp) -| (insert.north);

  \draw[<-,thick] (sample) -- ++(-1,0) node[above,align=center,rotate=90] (key) {Random\\key};

\end{tikzpicture}%

  }
  \caption{High-level representation of the proposed surrogate model and the computation of the reward for each sample path candidate. Each trapezoid represents a neural network module with learnable parameters, $\boldsymbol{\theta}$, which are different for each module. A different random key is used to sample each path candidate and each object within the path candidate, ensuring diversity in the generated paths.}
  \label{fig:model}
\end{figure}

We illustrate the high-level architecture of the proposed system in \cref{fig:model}. The pipeline first consists of a geometric transformation module that maps Cartesian coordinates of each scene object in $\mathbf{s}$ ($\boldsymbol{X}$) into a canonical frame ($\boldsymbol{X}'$) depending on the \gls{tx} and \gls{rx} positions. Next, an object encoder transforms each object in the input scene geometry into a latent representation, $\boldsymbol{Y}$. A DeepSets~\cite{deep-sets} architecture encodes the scene (i.e., all the objects) into a single feature vector, and a positional encoder generates a latent representation of the current path candidate. The \gls{gflownet} module combines the outputs of the three encoders to generate a flow, akin to a probability, for each object. The next object to be inserted into the path candidate is chosen at random, based on the probability distribution given by the output flows. A complete path candidate is generated by repeating the sampling sequence $K$ times, i.e., until an object is selected for each interaction. This path candidate is then fed to the path tracing module, e.g., using the image method~\cite{image-method} or minimization-based techniques~\cite{fpt,mpt}. Finally, in our model, the number of objects in the scene, $N$, can vary; however, each object is assumed to be represented by a fixed number of vertices, $V$ (e.g., the three vertices of a triangular face).

\subsection{Terms and Notation}

We use bold uppercase and lowercase letters (e.g., $\boldsymbol{X}$ or $\boldsymbol{x}$) to denote vectors, matrices or tensors, while regular letters (e.g., $X$ or $x$) represent scalars or specific physical entities, such as \glsxtrshort{tx} and \glsxtrshort{rx}. The $i$-th element (resp. row) of a vector (resp. matrix) $\boldsymbol{x}$ is denoted by $x_i$ (resp. $\boldsymbol{x}_i$). The value of a scalar $x$ or a tensor $\boldsymbol{x}$ at iteration $i$ is indicated by the superscript $x^{(i)}$ or $\boldsymbol{x}^{(i)}$. We denote the uniform distribution over the closed interval $[a,b]$ by $\mathcal{U}(a,b)$, where $u \sim \mathcal{U}(a,b)$ represents a random variable sampled from this distribution.

In the reinforcement learning framework, the \textbf{policy}, $\pi(a|s)$, is defined as a mapping from states to actions; specifically, it denotes the probability of taking an action, $a$, given the current state, $s$. Since an action $a$ and a current state $s$ fully determine the next state $s'$, the policy is sometimes written as $\pi(s'|s)$, to indicate the marginal probability of transitioning to a child state $s'$ given the parent state $s$. In our context, a state corresponds to a (partial) path candidate, $\boldsymbol{p}$, and an action corresponds to selecting the next object, $o_i \in \{o_1,...,o_N\}$, to extend the path. To indicate an incomplete path candidate, we denote it as
\begin{equation}
  \boldsymbol{p}=
  \begin{bmatrix}o_i & o_j & ... & -1
  \end{bmatrix},
\end{equation}
where each $o$ represents the index of an object in the scene and a $-1$ indicates that no object was chosen for the corresponding interaction yet.

The \textbf{reward function} $R(\boldsymbol{p})$ serves as an objective measure of the performance of the model. In the context of this study, the reward function is computed only for terminal states, i.e., when a path candidate is complete. It evaluates the physical validity of a sampled ray path and provides a non-zero scalar value when the sampling policy yields a candidate path that satisfies the underlying geometrical propagation constraints.

\subsection{Geometric Representation and Canonical Frame}
\label{subsec:geometric_transform}

To generalize across different scene configurations and \glsxtrlong{tx}-\glsxtrlong{rx} positions, we employ a geometric transformation that maps the scene into a canonical frame of reference. This preprocessing step ensures that the input features to the neural network are invariant to the global translation, rotation (azimuthal), and scaling of the scenario.

Let $\boldsymbol{x}_{\mathrm{\glsxtrshort{tx}}} \in \mathbb{R}^3$ and $\boldsymbol{x}_{\mathrm{\glsxtrshort{rx}}} \in \mathbb{R}^3$ denote the positions of the \glsxtrlong{tx} and \glsxtrlong{rx}, respectively. We define a local coordinate system basis $\mathcal{B} = \{\boldsymbol{u}, \boldsymbol{v}, \boldsymbol{w}\}$ and a scaling factor $s$ derived from the link geometry.

First, we define the longitudinal axis $\boldsymbol{w}$ (local $z$-axis) to be aligned with the line-of-sight vector from \gls{tx} to \gls{rx}, i.e.,
\begin{equation}
  \boldsymbol{w} = \frac{\boldsymbol{x}_{\mathrm{\glsxtrshort{rx}}} - \boldsymbol{x}_{\mathrm{\glsxtrshort{tx}}}}{s} \quad\text{with}\quad s = \|\boldsymbol{x}_{\mathrm{\glsxtrshort{rx}}} - \boldsymbol{x}_{\mathrm{\glsxtrshort{tx}}}\|.
\end{equation}

To fix the remaining axes while preserving the global vertical orientation, we define the lateral axis $\boldsymbol{u}$ (local $x$-axis) as orthogonal to both $\boldsymbol{w}$ and the global vertical axis $\boldsymbol{e}_z = [0, 0, 1]^\top$ by setting
\begin{equation}
  \boldsymbol{u} = \frac{\boldsymbol{w} \times \boldsymbol{e}_z}{\|\boldsymbol{w} \times \boldsymbol{e}_z\|}, \quad \boldsymbol{v} = \boldsymbol{w} \times \boldsymbol{u}.
\end{equation}

Here, $\boldsymbol{v}$ represents the local vertical axis. The basis matrix is given by $\boldsymbol{R} = [\boldsymbol{u}, \boldsymbol{v}, \boldsymbol{w}]^\top \in \mathbb{R}^{3 \times 3}$.

Given a vertex $\boldsymbol{x}_i$ of a scene object, the transformed coordinate $\boldsymbol{x}'_i$ is computed by translating the TX to the origin, normalizing by the link distance $s$, and projecting onto the basis $\mathcal{B}$, resulting in
\begin{equation} \label{eq:transform}
  \boldsymbol{x}'_i = \boldsymbol{R} \left( \frac{\boldsymbol{x}_i - \boldsymbol{x}_{\mathrm{\glsxtrshort{tx}}}}{s} \right).
\end{equation}

This transformation maps the \glsxtrlong{tx} to $(0,0,0)$ and the \glsxtrlong{rx} to $(0,0,1)$, creating a standardized representation regardless of the physical scale, absolute position, or orientation of the environment. While this transformation is only invariant to rotation around the vertical axis, this is sufficient for typical outdoor urban scenarios where the ground plane is approximately horizontal. We refer to \cref{appendix:invariance} for a formal proof of the invariance properties of this transformation.

\subsection{Generative Flow Networks for Path Sampling}

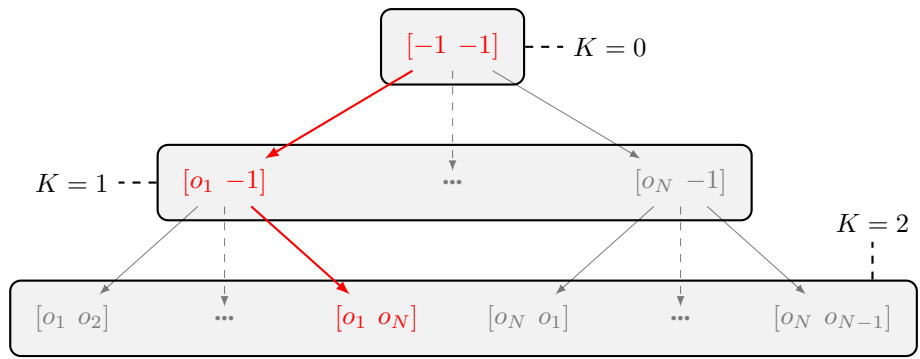
\begin{figure}[htbp]
  \centering
  \tikzsetnextfilename{tikzexternalize/path_candidates}%
  \begin{tikzpicture}
  [level distance=18mm,
    every node/.style={font=\bfseries, opacity=0.5},
    edge from parent/.style={draw, -latex, opacity=0.5},
    level 1/.style={sibling distance=30mm},
  level 2/.style={sibling distance=20mm}]
  \node[red, opacity=1] (k0) {$
    \begin{bmatrix}-1 & -1
  \end{bmatrix}$}
  child {node[red, opacity=1] (k11) {$
      \begin{bmatrix}o_1 & -1
    \end{bmatrix}$}
    child {node (k21) {$
        \begin{bmatrix}o_1 & o_2
      \end{bmatrix}$}
      edge from parent[black, thin]
    }
    child {node {...}
      edge from parent[dashed, black, thin]
    }
    child { node[red, opacity=1] {$
        \begin{bmatrix}o_1 & o_N
      \end{bmatrix}$}
      edge from parent[opacity=1, red, thick]
    }
    edge from parent[red, thick, opacity=1]
  }
  child {node {...}
    edge from parent[dashed]
  }
  child {node (k1n) {$
      \begin{bmatrix}o_N & - 1
    \end{bmatrix}$}
    child {node {$
        \begin{bmatrix}o_N & o_1
      \end{bmatrix}$}
      edge from parent
    }
    child {node {...}
      edge from parent[dashed]
    }
    child {node (k2n) {$
        \begin{bmatrix}o_N & o_{N-1}
      \end{bmatrix}$}
      edge from parent
    }
    edge from parent
  };
  \begin{pgfonlayer}{bg}
    \draw[rounded corners,thick,fill=gray!10] ([shift={(-.2,-.5)}]k0.west) rectangle ([shift={(.2,.5)}]k0.east) node[fitting node] (l0) {};
    \draw[rounded corners,thick,fill=gray!10] ([shift={(-.2,-.5)}]k11.west) rectangle ([shift={(.2,.5)}]k1n.east) node[fitting node] (l1) {};
    \draw[rounded corners,thick,fill=gray!10] ([shift={(-.2,-.5)}]k21.west) rectangle ([shift={(.2,.5)}]k2n.east) node[fitting node] (l2) {};
    \draw[dashed, opacity=1, thick] (l0.east) -- ++(+.5,0) node[anchor=west, opacity=1] {$K=0$};
    \draw[dashed, opacity=1, thick] (l1.west) -- ++(-.5,0) node[anchor=east, opacity=1] {$K=1$};
    \path (l2.north) -- (l2.north -| l2.east) node[pos=.9] (l2ne) {};
    \draw[dashed, opacity=1, thick] (l2ne.center) -- ++(0,.5) node[anchor=south, opacity=1] {$K=2$};
  \end{pgfonlayer}
\end{tikzpicture}%

  \caption{Illustration of the path generation process for second-order ray paths. Starting from the empty path candidate (\gls{tx}), the model sequentially selects scene objects from a forward policy $\pi(\boldsymbol{p}' | \boldsymbol{p})$. The process terminates after $K$ steps, and connects to \gls{rx}, forming a complete candidate path.}
  \label{fig:path_candidates}
\end{figure}

We formulate the ray sampling problem as a trajectory generation task on a directed tree, see \cref{fig:path_candidates}. Each state in the tree corresponds to a---possibly partial---path candidate. Taking a branch represents selecting the next object to extend the path. The root node implicitly represents the \gls{tx}, and the leaf nodes correspond to complete paths that terminate at the \gls{rx} after $K$ interactions.

The goal of the \gls{gflownet} model is to learn a forward policy $\pi(\boldsymbol{p}'|\boldsymbol{p})$ such that the marginal probability of sampling a complete path $\boldsymbol{p}$ is proportional to a given reward function $R(\boldsymbol{p})$~\cite{bengio2023gflownet}, i.e.,
\begin{equation}\label{eq:proportionality}
  P(\boldsymbol{p}) \propto R(\boldsymbol{p}).
\end{equation}

In the context of this paper, the reward $R(\boldsymbol{p})$ is simply defined as
\begin{equation}\label{eq:reward}
  R(\boldsymbol{p}) =
  \begin{cases}1, & \text{if } \boldsymbol{p} \text{ is a valid ray path}, \\ 0, & \text{otherwise}.
  \end{cases}
\end{equation}

A valid ray path is one that satisfies all geometrical constraints, such as visibility and reflection laws, as determined by the \textquote{Path Tracing} and \textquote{Geometry Validation} modules in \cref{fig:model}. This binary reward is intentional: it keeps the sampler independent of electromagnetic parameters such as frequency-dependent attenuation or material-specific losses. Instead of encoding those effects directly into the reward, we rely on geometry-aware priors elsewhere in the pipeline---most notably action masking and distance-based flow weighting (see hereafter)---to guide the model toward physically plausible candidates without tying it to a specific propagation regime.

To achieve the desired proportionality between path sampling probability and reward, the model must satisfy the following conditions:

\begin{enumerate}
  \item Each edge in the search graph must be assigned a positive flow, $F(\boldsymbol{p} \to \boldsymbol{p}') > 0$, where $\boldsymbol{p}$ is the parent state and $\boldsymbol{p}'$ is the child state;
  \item The total incoming flow to a state must equal the total outgoing flow plus the reward at that state, that is
    \begin{equation} \label{eq:flow_matching}
      \forall \boldsymbol{p}', F(\boldsymbol{p} \to \boldsymbol{p}') = R(\boldsymbol{p}') + \sum_{\boldsymbol{p}'' \in \mathcal{C}(\boldsymbol{p}')} F(\boldsymbol{p}' \to \boldsymbol{p}''),
    \end{equation}
    where $\mathcal{C}(\boldsymbol{p}')$ denotes the set of children of state $\boldsymbol{p}'$ in the search tree;
  \item The probability of selecting object $o_i$ given path candidate $\boldsymbol{p}$ must be defined as
    \begin{equation}
      \pi(o_i|\boldsymbol{p}) = \frac{F(\boldsymbol{p}, o_i)}{\sum_{j=1}^{N} F(\boldsymbol{p}, o_j)},
    \end{equation}
    that is, the probability of traversing an edge in the search graph is equal to its flow value, normalized over all outgoing edges.
\end{enumerate}

In practice, the forward policy is parameterized by a neural network with parameters, $\boldsymbol{\theta}$, and its output also depends on the encoded scene representation, $\boldsymbol{Y}$ (see \cref{fig:model}), but this dependency was omitted here for clarity.

\subsection{Training Procedure}

We train our model by minimizing the \gls{gflownet} loss function, which reformulates the flow matching constraint \eqref{eq:flow_matching} as the mean squared error
\begin{equation} \label{eq:loss}
  \mathcal{L}(\boldsymbol{p}') = \Big( F(\boldsymbol{p} \to \boldsymbol{p}') - R(\boldsymbol{p}') - \sum_{\boldsymbol{p}'' \in \mathcal{C}(\boldsymbol{p}')} F(\boldsymbol{p}' \to \boldsymbol{p}'') \Big)^2,
\end{equation}
which is accumulated over all visited states $\boldsymbol{p}'$ within the sampled trajectories.

The training procedure follows a standard gradient descent approach: at each iteration, we sample a batch of scenes and generate one or more path candidates per scene. For each sample, the geometric transformation (see \cref{subsec:geometric_transform}) is computed once, but the forward policy is evaluated $K$ times to sequentially select objects and construct a complete path candidate. The loss is accumulated across all intermediate states within each trajectory and across all candidates in the batch. Crucially, all model parameters---including those of the object encoder, scene encoder, positional encoder, and \gls{gflownet} module---are trained jointly via a single gradient descent step per batch.

\begin{figure}[htbp]
  \centering
  % TODO: do not resize, make tikz figure fit page width directly
  \resizebox{\textwidth}{!}{
  \tikzsetnextfilename{tikzexternalize/training_procedure}%
  \import{tikz}{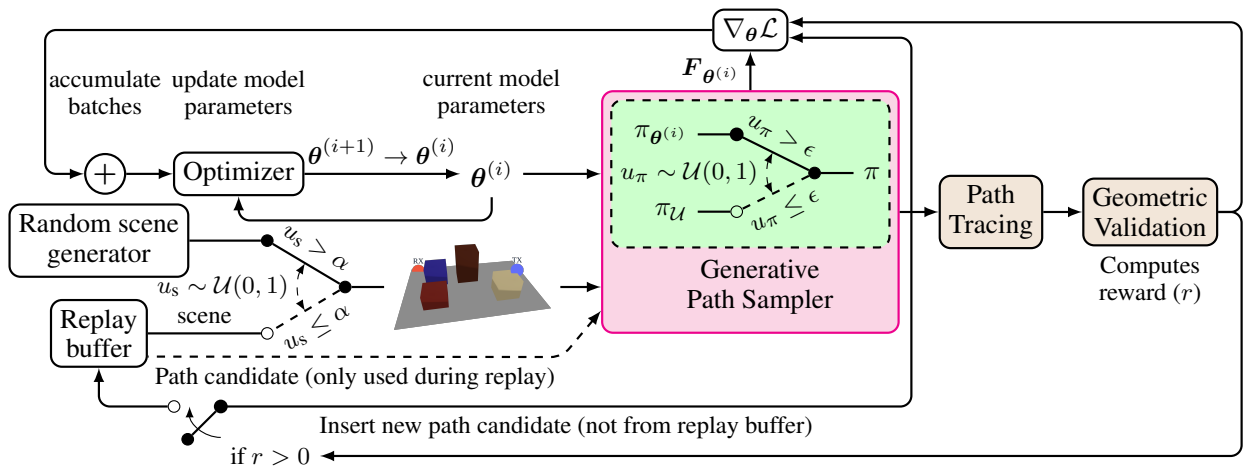}%

  }
  \caption{Training procedure for the \gls{gflownet}-based path sampler. At each iteration, a batch of path candidates is generated from a scene chosen either randomly or from a replay buffer. Each object in a path candidate is sampled using either the current flow-based policy, $\pi_{\boldsymbol{\theta}^{(i)}}$, or a uniform policy, $\pi_{\mathcal{U}}$. When a scene--path pair is replayed, the sampler regenerates the candidate with updated flow values. Each candidate is evaluated using the differentiable path tracing module to compute its reward. The gradient of the loss is calculated based on the generated paths, corresponding flows ($\boldsymbol{F}_{\boldsymbol{\theta}^{(i)}}$), and their rewards, followed by a model parameter update. If a new candidate yields a positive reward, it is stored in the replay buffer. The parameters $\alpha$ and $\epsilon$ control the probabilities of sampling from the replay buffer and using the uniform exploratory policy, respectively.}
  \label{fig:training_procedure}
\end{figure}

To enhance training performance and stability, we implement the architectural strategies detailed in the following subsections (see \cref{fig:training_procedure}).

\subsubsection{Successful Experience Replay Buffer}

In reinforcement learning, sparse rewards present a significant challenge because the model rarely receives the positive feedback necessary to associate actions with desired outcomes. This difficulty is particularly pronounced in our specific problem, where valid propagation paths represent only a tiny fraction of the candidate space. As detailed in \cref{sec:application}, the probability of randomly sampling a valid path decreases exponentially with the interaction order $K$ and scene size. Consequently, relying solely on random exploration can lead to training stagnation or convergence to trivial solutions, such as the model learning to set all flows to zero, also referred to as \emph{collapsing}.

To mitigate this, we maintain a \emph{successful experience replay buffer}. This buffer stores scene--path candidate pairs that have yielded positive rewards. During each training iteration, the model samples a scene from this buffer with probability $\alpha$ instead of generating a new configuration. This allows the framework to revisit and reinforce learning from successful trajectories, significantly improving convergence and training stability.

\subsubsection{Uniform Exploratory Policy}

To prevent the model from overfitting to its current policy and to encourage the discovery of diverse paths, we introduce an $\epsilon$-greedy strategy. With probability $\epsilon$, the next object in the sequence is sampled uniformly from the valid objects; with probability $1-\epsilon$, it is sampled according to the learned flow-based policy $\pi_{\boldsymbol{\theta}}$. This approach ensures persistent exploration even as the model converges toward an optimal sampling distribution.

\subsubsection{Action Masking}

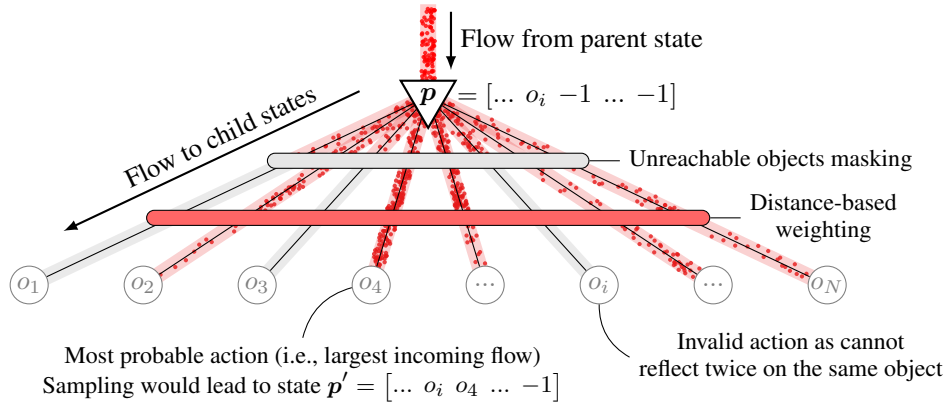
\begin{figure}[htbp]
  \centering
  \tikzsetnextfilename{tikzexternalize/action_pruning}%
  \begin{tikzpicture}
  [level distance=25mm,
  level 1/.style={sibling distance=15mm},]

  \newcommand{\drawFlow}[4]{
    % Draw the thick background line
    \begin{pgfonlayer}{bg}
      \draw[line width=6pt, opacity=0.2, #3] (#1.center) -- (#2.center);

      % Generate random points
      \ifnum#4>0
      \foreach \i in {1,...,#4}{
        % MATH EXPLANATION:
        % ($ ... $) : Calculation wrapper
        % (#1.center) ! .05 + 0.7*rnd ! (#2.center) : Pick a random point 15% to 95% along the line (avoids overlap with nodes)
        % ! 2.5pt * (rand) ! 90:(#2.center) : Move perpendicularly (90 deg) by a random amount between -2.5pt and 2.5pt
        \fill[#3, opacity=0.8]
        ($ (#1.center) ! .05 + 0.9*rnd ! (#2.center) ! 2.5pt * (rand) ! 90:(#2.center) $)
        circle (0.8pt);
      }
      \fi
    \end{pgfonlayer}
  }

  \node[draw, inner sep=1pt, regular polygon, regular polygon sides=3, shape border rotate=180, fill=white, thick, label=right:{$=
      \begin{bmatrix}... & o_i & -1 & ... & -1
  \end{bmatrix}$}] (sk) {$\boldsymbol{p}$}
  child {node[draw=black!50, text=black!50, circle, fill=white, minimum size=5mm, inner sep=0pt] (o1) {$o_1$}}
  child {node[draw=black!50, text=black!50, circle, fill=white, minimum size=5mm, inner sep=0pt] (o2) {$o_2$}}
  child {node[draw=black!50, text=black!50, circle, fill=white, minimum size=5mm, inner sep=0pt] (o3) {$o_3$}}
  child {node[draw=black!50, text=black!50, circle, fill=white, minimum size=5mm, inner sep=0pt] (o4) {$o_4$}}
  child {node[draw=black!50, text=black!50, circle, fill=white, minimum size=5mm, inner sep=0pt] (o5) {$...$}}
  child {node[draw=black!50, text=black!50, circle, fill=white, minimum size=5mm, inner sep=0pt] (oi) {$o_i$}}
  child {node[draw=black!50, text=black!50, circle, fill=white, minimum size=5mm, inner sep=0pt] (o7) {$...$}}
  child {node[draw=black!50, text=black!50, circle, fill=white, minimum size=5mm, inner sep=0pt] (on) {$o_N$}}
  ;

  \path (sk) -- (o1) node[pos=.33] (o1a) {} node[pos=.66] (o1b) {};
  \path (sk) -- (o2) node[pos=.33] (o2a) {} node[pos=.66] (o2b) {};
  \path (sk) -- (o3) node[pos=.33] (o3a) {} node[pos=.66] (o3b) {};
  \path (sk) -- (o4) node[pos=.33] (o4a) {} node[pos=.66] (o4b) {};
  \path (sk) -- (o5) node[pos=.33] (o5a) {} node[pos=.66] (o5b) {};
  \path (sk) -- (oi) node[pos=.33] (oia) {} node[pos=.66] (oib) {};
  \path (sk) -- (o7) node[pos=.33] (o7a) {} node[pos=.66] (o7b) {};
  \path (sk) -- (on) node[pos=.33] (ona) {} node[pos=.66] (onb) {};

  \draw[rounded corners=1mm, fill=gray!20] ([shift={(-.2,-.1)}]o1a.west) rectangle ([shift={(.2,.1)}]ona.east) node[fitting node] (mask1) {};

  \draw[rounded corners=1mm, fill=red!60] ([shift={(-.2,-.1)}]o1b.west) rectangle ([shift={(.2,.1)}]onb.east) node[fitting node] (mask2) {};

  \pgfmathsetseed{42}
  \coordinate (sktop) at ([yshift=10mm]sk.north);
  \drawFlow{sk}{sktop}{red}{100};

  \drawFlow{sk}{o1a}{red!80!gray}{20};
  \drawFlow{sk}{o2a}{red!80!gray}{40};
  \drawFlow{sk}{o3a}{red!80!gray}{30};
  \drawFlow{sk}{o4a}{red!80!gray}{80};
  \drawFlow{sk}{o5a}{red!80!gray}{70};
  \drawFlow{sk}{oia}{red!80!gray}{50};
  \drawFlow{sk}{o7a}{red!80!gray}{10};
  \drawFlow{sk}{ona}{red!80!gray}{30};

  \drawFlow{o1a}{o1b}{gray!80}{0};
  \drawFlow{o2a}{o2b}{red!80!gray}{40};
  \drawFlow{o3a}{o3b}{gray!80}{0};
  \drawFlow{o4a}{o4b}{red!80!gray}{80};
  \drawFlow{o5a}{o5b}{red!80!gray}{70};
  \drawFlow{oia}{oib}{gray!80}{0};
  \drawFlow{o7a}{o7b}{red!80!gray}{10};
  \drawFlow{ona}{onb}{red!80!gray}{30};

  \drawFlow{o1b}{o1}{gray!80}{0};
  \drawFlow{o2b}{o2}{red!80!gray}{10};
  \drawFlow{o3b}{o3}{gray!80}{0};
  \drawFlow{o4b}{o4}{red!80!gray}{80};
  \drawFlow{o5b}{o5}{red!80!gray}{20};
  \drawFlow{oib}{oi}{gray!80}{0};
  \drawFlow{o7b}{o7}{red!80!gray}{40};
  \drawFlow{onb}{on}{red!80!gray}{20};

  \draw (oi.south) to[bend right] ++(.4,-0.7) node[anchor=west,align=center,font=\small] {Invalid action as cannot\\reflect twice on the same object};
  \draw (o4) to[bend right] ++(-.9,-.7) node[anchor=north,align=center,font=\small] {Most probable action (i.e., largest incoming flow)\\Sampling would lead to state $\boldsymbol{p}' =
    \begin{bmatrix}... & o_i & o_4 & ... & -1
  \end{bmatrix}$};
  \draw (mask1.east) -- ++(.4,0) node[anchor=west,align=center,font=\small] {Unreachable objects masking};
  \draw (mask2.east) -- ++(.4,0) node[anchor=west,align=center,font=\small] {Distance-based\\weighting};

  \draw[thick, ->,shorten >=0.5cm,shorten <=1.0cm] ([yshift=0.5cm]sk.center) -- +($(o1)-(sk)$) node[midway,above,sloped] {Flow to child states};
  \draw[thick, ->,shorten >=0.35cm,shorten <=0.1cm] ([xshift=0.3cm]sktop.center) -- +($(sk)-(sktop)$) node[pos=0.4,right] {Flow from parent state};
\end{tikzpicture}%

  \caption{Action pruning via masking of invalid next-object choices based on geometric constraints. At each step of path construction, only objects capable of leading to a valid reflection towards the receiver are considered. This significantly reduces the action space and guides the \gls{gflownet} toward physically viable paths. The point distribution illustrates the \textquote{flow density}.}
  \label{fig:action_pruning}
\end{figure}

To prevent the model from wasting computational resources on physically impossible paths, we employ action masking (see \cref{fig:action_pruning}). At each sampling step, we compute a visibility mask that restricts the available actions to objects that are visible from the current interaction point. This acts as a hard constraint on the forward policy, effectively pruning the search tree.

While this technique was primarily designed for efficient inference, it also benefits training by reducing the sampling of invalid paths. This increases the proportion of valid paths in each batch, thereby providing a stronger and more consistent learning signal.

\subsubsection{Distance-Based Flow Weighting}

To further improve the ability of the model to identify promising paths, we apply a distance-based weighting to the flows prior to sampling (see \cref{fig:action_pruning}). Specifically, we multiply each outgoing flow by a weight based on the Euclidean distance between the last interaction point and the next object. For the final object, the distance to the receiver is also taken into account.

We define the weight for each object $o_i$ as
\begin{equation}
  w_i = \frac{d_i^{-2}}{\sum\limits_{j=1}^N d_j^{-2}},
\end{equation}
where $d_i$ is the distance metric (either to the last interaction point or to the receiver) for each object, and the exponent emphasizes the relationship between distance and decreased received power.

\subsubsection[Enforcing TX-RX Symmetry]{Enforcing \glsxtrshort{tx}-\glsxtrshort{rx} Symmetry}

Another important consideration in our approach is the inherent symmetry between the \gls{tx} and \gls{rx} in radio propagation. Assuming reciprocal propagation conditions, the path from \gls{tx} to \gls{rx} is physically equivalent to the reversed path from \gls{rx} to \gls{tx}. However, imposing this symmetry property directly on the machine learning architecture is not trivial. Instead, we incorporate this property by augmenting the training data with reversed path candidates. In our experiments (see \cref{sec:ablation_study}), this technique did not significantly improve model performance, possibly because the network architecture already possesses sufficient generalization capabilities. % Discuss more in future work, about possibility to add a second loss, checking equal flows in reverse?

\section{Application to Radio Coverage Map Prediction in an Ideal Urban Street Canyon}\label{sec:application}

To demonstrate the practical effectiveness of our machine-learning-assisted ray tracing framework, we present a comprehensive application to radio coverage prediction within urban street canyon environments. This scenario represents a canonical challenge in radio propagation modeling, characterized by complex multipath propagation involving multiple reflections from various building facades and the ground plane.

Urban street canyons provide an ideal environment for evaluating our approach for several reasons: (1) they exhibit distinct geometric patterns that can be learned by machine-learning-based models, (2) they generate a significant number of valid ray paths through higher-order reflections, and (3) they represent realistic scenarios frequently encountered in urban wireless network planning.

Our implementation leverages the DiffeRT differentiable ray tracing library~\cite{Eert2505:DiffeRT} for geometric computations and path validation, utilizing Equinox for model definition~\cite{equinox} and Optax~\cite{optax} for the optimization process. As the framework is built on the JAX ecosystem~\cite{jax2018github}, it benefits from just-in-time compilation and can be executed on CPUs, GPUs, or TPUs without additional implementation effort.

\subsection{Training Set}

\begin{figure}
  \centering
  \includegraphics[width=.5\textwidth]{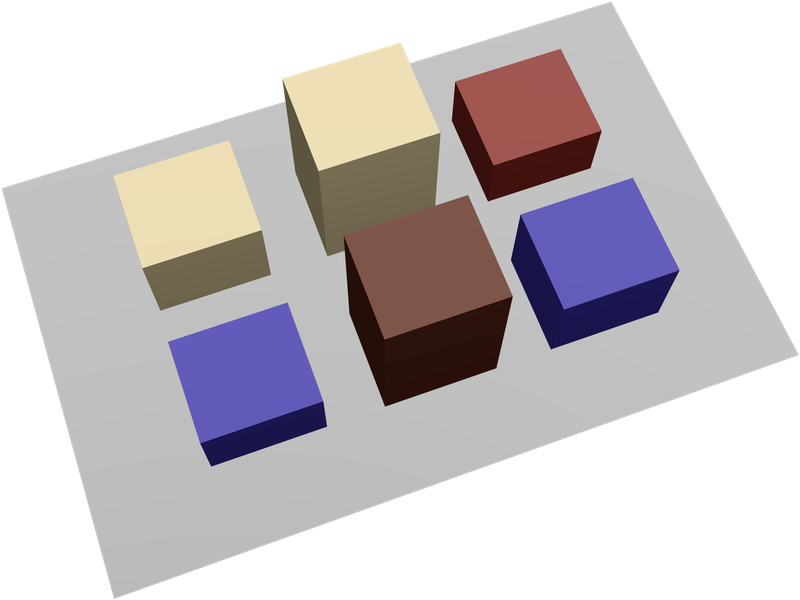}
  \caption{Street canyon scene from Sionna~RT~\cite{sionna-rt} used as the basis for generating our training and validation datasets.}
  \label{fig:base_scene}
\end{figure}

We utilize the urban street canyon scene provided by Sionna~RT~\cite{sionna-rt} as our base environment (see \cref{fig:base_scene}). This scene consists of a primary street flanked by buildings of varying heights, simulating the structural diversity of a realistic urban environment. The buildings are modeled as triangular mesh objects ($V=3$ in \cref{fig:model}).

To ensure robust generalization, we generate the training set by dynamically sampling modified versions of this base scene:
\begin{itemize}
  \item A random subset of buildings (ranging from \qtyrange{50}{100}{\percent}) is removed in each configuration; the inclusion of the ground plane can also be randomized (see \cref{tab:scene_statistics}).
  \item \gls{tx} and \gls{rx} positions are randomly sampled within two distinct regions: (a) exclusively within the street canyon and (b) across the entire scene. These variations ensure a broad range of propagation conditions, including challenging non-line-of-sight scenarios. The \gls{tx} height is randomized between \qty{2}{\meter} and \qty{50}{\meter}, while the \gls{rx} height is set between \qty{1}{\meter} and \qty{2}{\meter}.
\end{itemize}

Crucially, the training scenes are not static; a new random configuration is generated for every training iteration. This continuous variation exposes the model to a diverse architectural space, preventing overfitting to specific geometries. When sampling across the entire scene, the \gls{tx} or \gls{rx} may occasionally be placed inside a building. Although these represent degenerate scenarios, they are intentionally retained during training to force the model to learn the association between invalid geometries and zero rewards. Such cases are excluded only from the validation set to ensure an accurate assessment of model performance in physically plausible conditions.

\begin{table}[htbp]
  \centering
  \caption{Statistics from \num{100000} randomly generated scenes, showing the average number of unique path candidates and the percentage that are geometrically valid, for interaction orders $K\in\{0,1,2,3\}$. Values in \textbf{bold} denote configurations where the ground plane is always included; values in parentheses indicate scenarios where the ground plane is randomly included or excluded. Note the dramatic decrease in valid-path percentages for $K=2$ and especially $K=3$: higher-order reflections are intrinsically rarer because each additional reflection order introduces additional geometric constraints that most candidate paths fail to satisfy. The percentage of valid paths is consistently higher within the canyon region than across the whole scene (except marginally for $K=3$), reflecting the more constrained and challenging non-line-of-sight configurations in the full scene.}
  \label{tab:scene_statistics}
  % https://stackoverflow.com/a/69290979
  \sisetup{
    text-series-to-math = true,
    propagate-math-font = true
  }
  \begin{tabular}{@{}lccc@{}}
    \toprule
    &
    \begin{tabular}{@{}c@{}}Average number of\\path candidates
    \end{tabular} & \multicolumn{2}{c}{\% of valid path candidates} \\ \midrule
    \begin{tabular}{@{}c@{}}Sampling\\region
    \end{tabular} &  & Canyon & Whole scene \\\cmidrule{1-1}
    K & & & \\
    0 & \textbf{1} (1) & \textbf{100} (100) & \textbf{33.6} (33.6) \\
    1 & \textbf{56} (55) & \textbf{\num{3.66}} (\num{3.05}) & \textbf{\num{1.43}} (\num{1.14}) \\
    2 & \textbf{\num{3333}} (\num{3287}) & \textbf{\num{4.15e-2}} (\num{3.28e-2}) & \textbf{\num{1.94e-2}} (\num{1.52e-2}) \\
    3 & \textbf{\num{208750}} (\num{205064}) & \textbf{\num{2.6e-4}} (\num{2.19e-3}) & \textbf{\num{3.27e-4}} (\num{3.00e-4}) \\ \bottomrule
  \end{tabular}
\end{table}

\subsection{Simulation Parameters}

Simulations were conducted across various street canyon configurations, with models trained for interaction orders ranging from $K=1$ to $K=3$. Each training iteration consists of sampling a batch of $B=64$ path candidates from a dynamically generated scene. When the replay buffer is active, an additional batch of $B=64$ trajectories is drawn from the buffer to reinforce previous successful samples.

Unless otherwise specified, the architectural and hyperparameter configurations used throughout our experiments are as follows.
\begin{itemize}
  \item \textbf{Embedding Size:} The dimensionality of the latent embeddings throughout the model is set to $d=128$.
  \item \textbf{Object Encoder:} Each of the $N$ triangular facets is encoded via an \gls{mlp} consisting of two hidden layers with $2d$ units each and ReLU activations. The output layer has size $d$, resulting in $N$ distinct object embeddings.
  \item \textbf{State Encoder:} The current state (i.e., the partial path candidate) is encoded through a single linear layer. It accepts the object embeddings corresponding to the currently selected sequence and produces a state vector of size $d'=K \cdot d$.
  \item \textbf{Scene Encoder:} A global scene representation is computed by averaging all object embeddings and passing the result through an \gls{mlp} with two hidden layers of $2d$ units, ReLU activations, and an output dimensionality of $d''=d$.
  \item \textbf{Flow Computation:} Individual flows are computed by an \gls{mlp} with two hidden layers of $2(d+d'+d'')$ units and LeakyReLU activations. The network produces $N$ scalar outputs, to which an exponential function is applied to ensure strictly positive flow values.
  \item \textbf{Replay Buffer:} When enabled, the buffer capacity is set to \num{10000} paths with a weighting factor of $\alpha=0.5$. The weighting factor balances the contribution of newly sampled paths and those drawn from the buffer during training.
  \item \textbf{Exploratory Policy:} The $\epsilon$-greedy exploratory strategy utilizes a fixed probability of $\epsilon=0.1$. A decreasing schedule for $\epsilon$ was also considered, but it did not produce better convergence results than a fixed value.
\end{itemize}

Training is performed using the Muon optimizer~\cite{jordan2024muon} with a learning rate of \num{1e-4}. Each model is trained for \num{500000} iterations, which was found to be sufficient for convergence across all tested configurations. Performance is monitored every \num{1000} iterations by evaluating the model on a validation set of \num{100} randomly generated scenes. The hit rate during validation is estimated by drawing $M=10$ samples per scene.

\subsection{Results}

We evaluate our machine-learning-assisted ray tracing framework across three key dimensions: (1) the impact of individual architectural and algorithmic components through a systematic ablation study, from the convergence characteristics of performance metrics observed during training, (2) the computational efficiency compared to exhaustive ray tracing, and (3) the accuracy of predicted coverage maps for the ultimate application task. Together, these evaluations demonstrate both the practical effectiveness and computational advantages of our approach for radio propagation modeling in complex urban environments.

\subsubsection{Performance Metrics}\label{sec:performance_metrics}

To evaluate the effectiveness of the proposed generative framework, we employ two complementary metrics that quantify both the precision and the coverage of the path sampler:

\begin{itemize}
  \item \textbf{Accuracy:} This metric represents the ratio of valid ray paths to the total number of sampled path candidates. It serves as an empirical estimate of the probability of sampling a valid path, as defined in \eqref{eq:proportionality}. An accuracy of \qty{100}{\percent} signifies that the model has successfully learned to exclusively sample candidates that return a non-zero reward.
  \item \textbf{Hit Rate:} Defined as the fraction of all unique valid ray paths identified by the sampler, this metric measures the ability of the model to recover the complete set of valid propagation paths. A hit rate of \qty{100}{\percent} indicates that the sampler has successfully discovered every possible solution within the scene.
\end{itemize}

While the hit rate provides a definitive measure of the sampling coverage, its calculation requires an exhaustive ground-truth search to identify all potential paths in advance. Furthermore, the hit rate is intrinsically linked to the number of samples $M$ drawn from the model. Consequently, we utilize the hit rate primarily for validation and benchmarking. In contrast, accuracy can be computed both during training and during inference without prior scene knowledge and remains independent of the sampling budget $M$. However, accuracy alone does not account for the diversity of the generated paths; thus, both metrics are necessary to fully characterize the performance of the model.

\subsubsection{Ablation Study}\label{sec:ablation_study}

To understand the contribution of each component in our framework, we conducted a systematic ablation study. Specifically, we isolate and evaluate the impact of key architectural choices and algorithmic techniques introduced in \cref{sec:methodology}, including the replay buffer, exploratory policy, action masking, distance-based flow weighting, and symmetry enforcement. Additionally, we examine the importance of training scenario diversity. For each ablation, we train models under controlled conditions, varying only the component under investigation while keeping all other hyperparameters fixed. We discuss the effectiveness of the proposed methods in \cref{sec:framework_effectiveness}.

\paragraph{Replay Buffer and Exploratory Policy}

\begin{figure}
  \centering
  \tikzsetnextfilename{tikzexternalize/ablation_buffer_epsilon}%
  \begin{tikzpicture}[every plot/.append style={mark repeat=30},every mark/.append style={opacity=0.7}]
  \newcommand{\plotwidth}{5.4cm}
  \newcommand{\plotheight}{4cm}
  \pgfplotscreateplotcyclelist{custom cycle list}{
    blue,every mark/.append style={fill=blue!80!black},mark=*\\
    red!20!black,every mark/.append style={fill=red!80!black},mark=otimes*\\
    brown,every mark/.append style={fill=brown!80!black},mark=square*\\
    gray,every mark/.append style={fill=gray!80!black},mark=diamond*\\
  }
  \pgfplotsset{every axis/.append style={scaled x ticks=false}}
  \pgfplotsset{every x tick label/.append style={/pgf/number format/fixed}}

  % Accuracy

  \begin{axis}[
      name=plot_order_1_sr,
      width=\plotwidth,
      height=\plotheight,
      ylabel={Accuracy (\%)},
      title={$K=1$},
      xtick pos=left,
      xtick={0,1.25e5, 2.5e5, 3.75e5, 5e5},
      xticklabels=\empty,
      ymin=0,ymax=100,
      ymajorgrids=true,
      cycle list name=custom cycle list,
    ]
    \addplot table [x=episode,y=success_rate,col sep=comma] {data/replay_and_exploratory_policy/o1_e0_r0.txt};
    \addplot table [x=episode,y=success_rate,col sep=comma] {data/replay_and_exploratory_policy/o1_e0_r1.txt};
    \addplot table [x=episode,y=success_rate,col sep=comma] {data/replay_and_exploratory_policy/o1_e1_r0.txt};
    \addplot table [x=episode,y=success_rate,col sep=comma] {data/replay_and_exploratory_policy/o1_e1_r1.txt};
  \end{axis}

  \begin{axis}[
      name=plot_order_2_sr,
      at={($ (.1cm,0cm) + (plot_order_1_sr.south east) $)},
      width=\plotwidth,
      height=\plotheight,
      title={$K=2$},
      title style={name=title},
      xtick pos=left,
      xtick={0,1.25e5, 2.5e5, 3.75e5, 5e5},
      xticklabels=\empty,
      yticklabels=\empty,
      ymin=0,ymax=100,
      ymajorgrids=true,
      cycle list name=custom cycle list,
      legend style={font=\small, at={([yshift=1ex]title.north)},anchor=south,name=leg,nodes={anchor=base}},
      legend columns=-1,legend style={column sep=1ex},
      legend image post style={yshift=.5ex},
      every legend to name picture/.style={
        baseline={(leg.base)},
      }
    ]
    \addlegendimage{blue,only marks,fill=blue!80!black,mark=*}
    \addlegendimage{red!20!black,only marks,fill=red!80!black,mark=otimes*}
    \addlegendimage{brown,only marks,fill=brown!80,mark=square*}
    \addlegendimage{gray,only marks,fill=gray!80!black,mark=diamond*}
    \legend{baseline,replay,$\epsilon$,replay + $\epsilon$}
    \addplot table [x=episode,y=success_rate,col sep=comma] {data/replay_and_exploratory_policy/o2_e0_r0.txt};
    \addplot table [x=episode,y=success_rate,col sep=comma] {data/replay_and_exploratory_policy/o2_e0_r1.txt};
    \addplot table [x=episode,y=success_rate,col sep=comma] {data/replay_and_exploratory_policy/o2_e1_r0.txt};
    \addplot table [x=episode,y=success_rate,col sep=comma] {data/replay_and_exploratory_policy/o2_e1_r1.txt};
  \end{axis}

  \begin{axis}[
      name=plot_order_3_sr,
      at={($ (.1cm,0cm) + (plot_order_2_sr.south east) $)},
      width=\plotwidth,
      height=\plotheight,
      title={$K=3$},
      xtick pos=left,
      xtick={0,1.25e5, 2.5e5, 3.75e5, 5e5},
      xticklabels=\empty,
      yticklabels=\empty,
      ymin=0,ymax=100,
      ymajorgrids=true,
      cycle list name=custom cycle list,
    ]
    \addplot table [x=episode,y=success_rate,col sep=comma] {data/replay_and_exploratory_policy/o3_e0_r0.txt};
    \addplot table [x=episode,y=success_rate,col sep=comma] {data/replay_and_exploratory_policy/o3_e0_r1.txt};
    \addplot table [x=episode,y=success_rate,col sep=comma] {data/replay_and_exploratory_policy/o3_e1_r0.txt};
    \addplot table [x=episode,y=success_rate,col sep=comma] {data/replay_and_exploratory_policy/o3_e1_r1.txt};
  \end{axis}

  % Hit rate

  \begin{axis}[
      name=plot_order_1_hr,
      anchor=north,
      at={($ (0cm,-.3cm) + (plot_order_1_sr.south) $)},
      width=\plotwidth,
      height=\plotheight,
      ylabel={Hit rate (\%)},
      xtick pos=left,
      xtick={0,1.25e5, 2.5e5, 3.75e5, 5e5},
      xticklabels={0,,250k,,500k},
      x tick label style={/pgf/number format/fixed},
      ymin=0,ymax=100,
      ymajorgrids=true,
      cycle list name=custom cycle list,
    ]
    \addplot table [x=episode,y=hit_rate,col sep=comma] {data/replay_and_exploratory_policy/o1_e0_r0.txt};
    \addplot table [x=episode,y=hit_rate,col sep=comma] {data/replay_and_exploratory_policy/o1_e0_r1.txt};
    \addplot table [x=episode,y=hit_rate,col sep=comma] {data/replay_and_exploratory_policy/o1_e1_r0.txt};
    \addplot table [x=episode,y=hit_rate,col sep=comma] {data/replay_and_exploratory_policy/o1_e1_r1.txt};
  \end{axis}

  \begin{axis}[
      name=plot_order_2_hr,
      at={($ (.1cm,0cm) + (plot_order_1_hr.south east) $)},
      width=\plotwidth,
      height=\plotheight,
      xlabel={Training episodes},
      xtick pos=left,
      xtick={0,1.25e5, 2.5e5, 3.75e5, 5e5},
      xticklabels={0,,250k,,500k},
      yticklabels=\empty,
      ymin=0,ymax=100,
      ymajorgrids=true,
      cycle list name=custom cycle list,
    ]
    \addplot table [x=episode,y=hit_rate,col sep=comma] {data/replay_and_exploratory_policy/o2_e0_r0.txt};
    \addplot table [x=episode,y=hit_rate,col sep=comma] {data/replay_and_exploratory_policy/o2_e0_r1.txt};
    \addplot table [x=episode,y=hit_rate,col sep=comma] {data/replay_and_exploratory_policy/o2_e1_r0.txt};
    \addplot table [x=episode,y=hit_rate,col sep=comma] {data/replay_and_exploratory_policy/o2_e1_r1.txt};
  \end{axis}

  \begin{axis}[
      name=plot_order_3_hr,
      at={($ (.1cm,0cm) + (plot_order_2_hr.south east) $)},
      width=\plotwidth,
      height=\plotheight,
      xtick pos=left,
      xtick={0,1.25e5, 2.5e5, 3.75e5, 5e5},
      xticklabels={0,,250k,,500k},
      yticklabels=\empty,
      ymin=0,ymax=100,
      ymajorgrids=true,
      cycle list name=custom cycle list,
    ]
    \addplot table [x=episode,y=hit_rate,col sep=comma] {data/replay_and_exploratory_policy/o3_e0_r0.txt};
    \addplot table [x=episode,y=hit_rate,col sep=comma] {data/replay_and_exploratory_policy/o3_e0_r1.txt};
    \addplot table [x=episode,y=hit_rate,col sep=comma] {data/replay_and_exploratory_policy/o3_e1_r0.txt};
    \addplot table [x=episode,y=hit_rate,col sep=comma] {data/replay_and_exploratory_policy/o3_e1_r1.txt};
  \end{axis}
\end{tikzpicture}%

  \caption{Replay buffer and exploratory policy impact on performance metrics during training.}
  \label{fig:ablation_buffer_epsilon}
\end{figure}
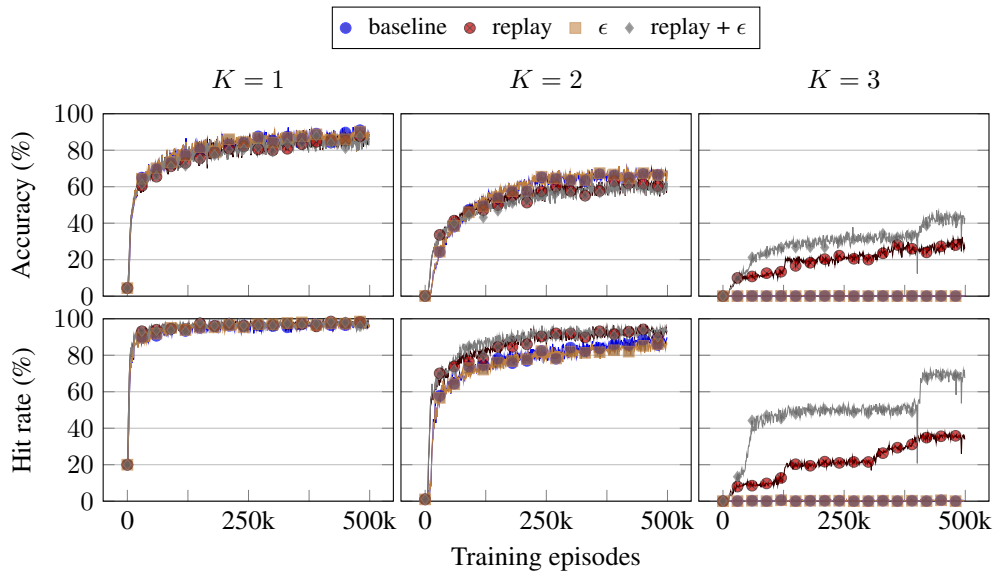

The replay buffer and exploratory policy are critical mechanisms for balancing exploration and exploitation during training. The replay buffer maintains a history of successfully discovered paths to ensure that the model does not collapse. It does so by enforcing that a fixed proportion of generated paths must be valid. Meanwhile, the $\epsilon$-greedy exploratory policy ensures sufficient exploration of the action space to discover new valid paths. We evaluate models trained with and without these components across different interaction orders to quantify their impact on convergence speed and final hit rate performance. \Cref{fig:ablation_buffer_epsilon} shows the training curves for these configurations and illustrates how each component contributes to overall performance.

\paragraph{Action Masking and Distance-Based Weighting}

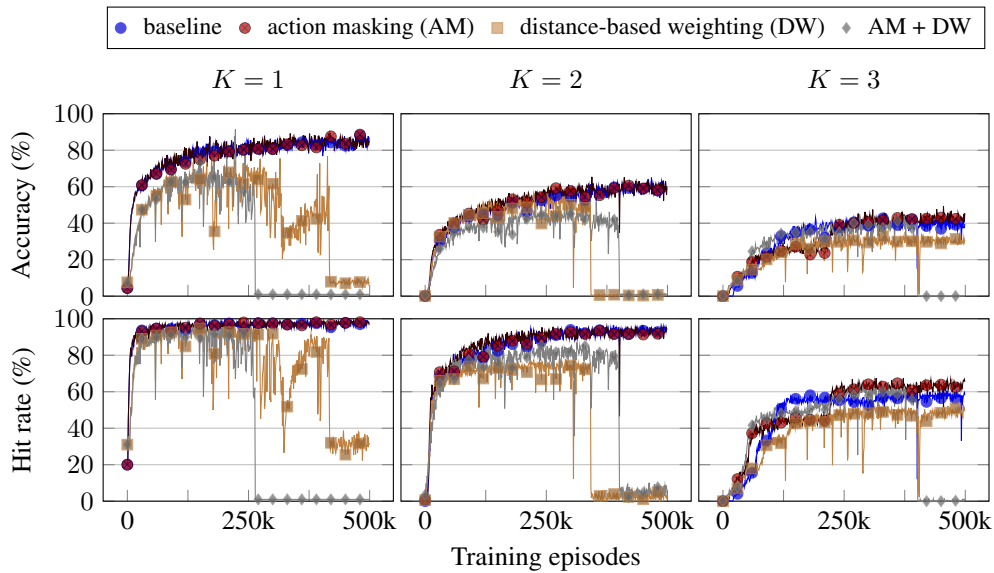
\begin{figure}
  \centering
  \tikzsetnextfilename{tikzexternalize/ablation_masking_weighting}%
  \begin{tikzpicture}[every plot/.append style={mark repeat=30},every mark/.append style={opacity=0.7}]
  \newcommand{\plotwidth}{5.4cm}
  \newcommand{\plotheight}{4cm}
  \pgfplotscreateplotcyclelist{custom cycle list}{
    blue,every mark/.append style={fill=blue!80!black},mark=*\\
    red!20!black,every mark/.append style={fill=red!80!black},mark=otimes*\\
    brown,every mark/.append style={fill=brown!80!black},mark=square*\\
    gray,every mark/.append style={fill=gray!80!black},mark=diamond*\\
  }
  \pgfplotsset{every axis/.append style={scaled x ticks=false}}
  \pgfplotsset{every x tick label/.append style={/pgf/number format/fixed}}

  % Accuracy

  \begin{axis}[
      name=plot_order_1_sr,
      width=\plotwidth,
      height=\plotheight,
      ylabel={Accuracy (\%)},
      title={$K=1$},
      xtick pos=left,
      xtick={0,1.25e5, 2.5e5, 3.75e5, 5e5},
      xticklabels=\empty,
      ymin=0,ymax=100,
      ymajorgrids=true,
      cycle list name=custom cycle list,
    ]
    \addplot table [x=episode,y=success_rate,col sep=comma] {data/action_masking_and_weighting/o1_am0_dw0.txt};
    \addplot table [x=episode,y=success_rate,col sep=comma] {data/action_masking_and_weighting/o1_am1_dw0.txt};
    \addplot table [x=episode,y=success_rate,col sep=comma] {data/action_masking_and_weighting/o1_am0_dw1.txt};
    \addplot table [x=episode,y=success_rate,col sep=comma] {data/action_masking_and_weighting/o1_am1_dw1.txt};
  \end{axis}

  \begin{axis}[
      name=plot_order_2_sr,
      at={($ (.1cm,0cm) + (plot_order_1_sr.south east) $)},
      width=\plotwidth,
      height=\plotheight,
      title={$K=2$},
      title style={name=title},
      xtick pos=left,
      xtick={0,1.25e5, 2.5e5, 3.75e5, 5e5},
      xticklabels=\empty,
      yticklabels=\empty,
      ymin=0,ymax=100,
      ymajorgrids=true,
      cycle list name=custom cycle list,
      legend style={font=\small, at={([yshift=1ex]title.north)},anchor=south,name=leg,nodes={anchor=base}},
      legend columns=-1,legend style={column sep=1ex},
      legend image post style={yshift=.5ex},
      every legend to name picture/.style={
        baseline={(leg.base)},
      }
    ]
    \addlegendimage{blue,only marks,fill=blue!80!black,mark=*}
    \addlegendimage{red!20!black,only marks,fill=red!80!black,mark=otimes*}
    \addlegendimage{brown,only marks,fill=brown!80,mark=square*}
    \addlegendimage{gray,only marks,fill=gray!80!black,mark=diamond*}
    \legend{baseline,action masking (AM),distance-based weighting (DW),AM + DW}
    \addplot table [x=episode,y=success_rate,col sep=comma] {data/action_masking_and_weighting/o2_am0_dw0.txt};
    \addplot table [x=episode,y=success_rate,col sep=comma] {data/action_masking_and_weighting/o2_am1_dw0.txt};
    \addplot table [x=episode,y=success_rate,col sep=comma] {data/action_masking_and_weighting/o2_am0_dw1.txt};
    \addplot table [x=episode,y=success_rate,col sep=comma] {data/action_masking_and_weighting/o2_am1_dw1.txt};
  \end{axis}

  \begin{axis}[
      name=plot_order_3_sr,
      at={($ (.1cm,0cm) + (plot_order_2_sr.south east) $)},
      width=\plotwidth,
      height=\plotheight,
      title={$K=3$},
      xtick pos=left,
      xtick={0,1.25e5, 2.5e5, 3.75e5, 5e5},
      xticklabels=\empty,
      yticklabels=\empty,
      ymin=0,ymax=100,
      ymajorgrids=true,
      cycle list name=custom cycle list,
    ]
    \addplot table [x=episode,y=success_rate,col sep=comma] {data/action_masking_and_weighting/o3_am0_dw0.txt};
    \addplot table [x=episode,y=success_rate,col sep=comma] {data/action_masking_and_weighting/o3_am1_dw0.txt};
    \addplot table [x=episode,y=success_rate,col sep=comma] {data/action_masking_and_weighting/o3_am0_dw1.txt};
    \addplot table [x=episode,y=success_rate,col sep=comma] {data/action_masking_and_weighting/o3_am1_dw1.txt};
  \end{axis}

  % Hit rate

  \begin{axis}[
      name=plot_order_1_hr,
      anchor=north,
      at={($ (0cm,-.3cm) + (plot_order_1_sr.south) $)},
      width=\plotwidth,
      height=\plotheight,
      ylabel={Hit rate (\%)},
      xtick pos=left,
      xtick={0,1.25e5, 2.5e5, 3.75e5, 5e5},
      xticklabels={0,,250k,,500k},
      x tick label style={/pgf/number format/fixed},
      ymin=0,ymax=100,
      ymajorgrids=true,
      cycle list name=custom cycle list,
    ]
    \addplot table [x=episode,y=hit_rate,col sep=comma] {data/action_masking_and_weighting/o1_am0_dw0.txt};
    \addplot table [x=episode,y=hit_rate,col sep=comma] {data/action_masking_and_weighting/o1_am1_dw0.txt};
    \addplot table [x=episode,y=hit_rate,col sep=comma] {data/action_masking_and_weighting/o1_am0_dw1.txt};
    \addplot table [x=episode,y=hit_rate,col sep=comma] {data/action_masking_and_weighting/o1_am1_dw1.txt};
  \end{axis}

  \begin{axis}[
      name=plot_order_2_hr,
      at={($ (.1cm,0cm) + (plot_order_1_hr.south east) $)},
      width=\plotwidth,
      height=\plotheight,
      xlabel={Training episodes},
      xtick pos=left,
      xtick={0,1.25e5, 2.5e5, 3.75e5, 5e5},
      xticklabels={0,,250k,,500k},
      yticklabels=\empty,
      ymin=0,ymax=100,
      ymajorgrids=true,
      cycle list name=custom cycle list,
    ]
    \addplot table [x=episode,y=hit_rate,col sep=comma] {data/action_masking_and_weighting/o2_am0_dw0.txt};
    \addplot table [x=episode,y=hit_rate,col sep=comma] {data/action_masking_and_weighting/o2_am1_dw0.txt};
    \addplot table [x=episode,y=hit_rate,col sep=comma] {data/action_masking_and_weighting/o2_am0_dw1.txt};
    \addplot table [x=episode,y=hit_rate,col sep=comma] {data/action_masking_and_weighting/o2_am1_dw1.txt};
  \end{axis}

  \begin{axis}[
      name=plot_order_3_hr,
      at={($ (.1cm,0cm) + (plot_order_2_hr.south east) $)},
      width=\plotwidth,
      height=\plotheight,
      xtick pos=left,
      xtick={0,1.25e5, 2.5e5, 3.75e5, 5e5},
      xticklabels={0,,250k,,500k},
      yticklabels=\empty,
      ymin=0,ymax=100,
      ymajorgrids=true,
      cycle list name=custom cycle list,
    ]
    \addplot table [x=episode,y=hit_rate,col sep=comma] {data/action_masking_and_weighting/o3_am0_dw0.txt};
    \addplot table [x=episode,y=hit_rate,col sep=comma] {data/action_masking_and_weighting/o3_am1_dw0.txt};
    \addplot table [x=episode,y=hit_rate,col sep=comma] {data/action_masking_and_weighting/o3_am0_dw1.txt};
    \addplot table [x=episode,y=hit_rate,col sep=comma] {data/action_masking_and_weighting/o3_am1_dw1.txt};
  \end{axis}
\end{tikzpicture}%

  \caption{Action masking and distance-based weighting impact on performance metrics during training.}
  \label{fig:ablation_masking_weighting}
\end{figure}

Action masking prevents the model from selecting physically invalid actions. For example, it prevents consecutive interactions with the same object or objects \textquote{behind} the current reflecting interface (for second-order interactions and higher). Distance-based flow weighting biases the sampling distribution toward geometrically shorter paths by down-weighting objects that are far from the current ray direction. These techniques encode domain knowledge directly into the sampling process. We investigate whether these heuristics improve sample efficiency and final performance, or whether the model can learn equivalent behavior from data through sufficient training iterations alone. \Cref{fig:ablation_masking_weighting} presents the training curves for these configurations, illustrating how each component contributes to overall performance.

\paragraph{Symmetry Enforcement}

\begin{figure}
  \centering
  \tikzsetnextfilename{tikzexternalize/ablation_symmetry}%
  \begin{tikzpicture}[every plot/.append style={mark repeat=30},every mark/.append style={opacity=0.7}]
  \newcommand{\plotwidth}{5.4cm}
  \newcommand{\plotheight}{4cm}
  \pgfplotscreateplotcyclelist{custom cycle list}{
    blue,every mark/.append style={fill=blue!80!black},mark=*\\
    red!20!black,every mark/.append style={fill=red!80!black},mark=otimes*\\
  }
  \pgfplotsset{every axis/.append style={scaled x ticks=false}}
  \pgfplotsset{every x tick label/.append style={/pgf/number format/fixed}}

  % Accuracy

  \begin{axis}[
      name=plot_order_1_sr,
      width=\plotwidth,
      height=\plotheight,
      ylabel={Accuracy (\%)},
      title={$K=1$},
      xtick pos=left,
      xtick={0,1.25e5, 2.5e5, 3.75e5, 5e5},
      xticklabels=\empty,
      ymin=0,ymax=100,
      ymajorgrids=true,
      cycle list name=custom cycle list,
    ]
    \addplot table [x=episode,y=success_rate,col sep=comma] {data/symmetry/o1_sym0.txt};
    \addplot table [x=episode,y=success_rate,col sep=comma] {data/symmetry/o1_sym1.txt};
  \end{axis}

  \begin{axis}[
      name=plot_order_2_sr,
      at={($ (.1cm,0cm) + (plot_order_1_sr.south east) $)},
      width=\plotwidth,
      height=\plotheight,
      title={$K=2$},
      title style={name=title},
      xtick pos=left,
      xtick={0,1.25e5, 2.5e5, 3.75e5, 5e5},
      xticklabels=\empty,
      yticklabels=\empty,
      ymin=0,ymax=100,
      ymajorgrids=true,
      cycle list name=custom cycle list,
      legend style={font=\small, at={([yshift=1ex]title.north)},anchor=south,name=leg,nodes={anchor=base}},
      legend columns=-1,legend style={column sep=1ex},
      legend image post style={yshift=.5ex},
      every legend to name picture/.style={
        baseline={(leg.base)},
      }
    ]
    \addlegendimage{blue,only marks,fill=blue!80!black,mark=*}
    \addlegendimage{red!20!black,only marks,fill=red!80!black,mark=otimes*}
    \legend{baseline,replaying symmetric scenarios}
    \addplot table [x=episode,y=success_rate,col sep=comma] {data/symmetry/o2_sym0.txt};
    \addplot table [x=episode,y=success_rate,col sep=comma] {data/symmetry/o2_sym1.txt};
  \end{axis}

  \begin{axis}[
      name=plot_order_3_sr,
      at={($ (.1cm,0cm) + (plot_order_2_sr.south east) $)},
      width=\plotwidth,
      height=\plotheight,
      title={$K=3$},
      xtick pos=left,
      xtick={0,1.25e5, 2.5e5, 3.75e5, 5e5},
      xticklabels=\empty,
      yticklabels=\empty,
      ymin=0,ymax=100,
      ymajorgrids=true,
      cycle list name=custom cycle list,
    ]
    \addplot table [x=episode,y=success_rate,col sep=comma] {data/symmetry/o3_sym0.txt};
    \addplot table [x=episode,y=success_rate,col sep=comma] {data/symmetry/o3_sym1.txt};
  \end{axis}

  % Hit rate

  \begin{axis}[
      name=plot_order_1_hr,
      anchor=north,
      at={($ (0cm,-.3cm) + (plot_order_1_sr.south) $)},
      width=\plotwidth,
      height=\plotheight,
      ylabel={Hit rate (\%)},
      xtick pos=left,
      xtick={0,1.25e5, 2.5e5, 3.75e5, 5e5},
      xticklabels={0,,250k,,500k},
      x tick label style={/pgf/number format/fixed},
      ymin=0,ymax=100,
      ymajorgrids=true,
      cycle list name=custom cycle list,
    ]
    \addplot table [x=episode,y=hit_rate,col sep=comma] {data/symmetry/o1_sym0.txt};
    \addplot table [x=episode,y=hit_rate,col sep=comma] {data/symmetry/o1_sym1.txt};
  \end{axis}

  \begin{axis}[
      name=plot_order_2_hr,
      at={($ (.1cm,0cm) + (plot_order_1_hr.south east) $)},
      width=\plotwidth,
      height=\plotheight,
      xlabel={Training episodes},
      xtick pos=left,
      xtick={0,1.25e5, 2.5e5, 3.75e5, 5e5},
      xticklabels={0,,250k,,500k},
      yticklabels=\empty,
      ymin=0,ymax=100,
      ymajorgrids=true,
      cycle list name=custom cycle list,
    ]
    \addplot table [x=episode,y=hit_rate,col sep=comma] {data/symmetry/o2_sym0.txt};
    \addplot table [x=episode,y=hit_rate,col sep=comma] {data/symmetry/o2_sym1.txt};
  \end{axis}

  \begin{axis}[
      name=plot_order_3_hr,
      at={($ (.1cm,0cm) + (plot_order_2_hr.south east) $)},
      width=\plotwidth,
      height=\plotheight,
      xtick pos=left,
      xtick={0,1.25e5, 2.5e5, 3.75e5, 5e5},
      xticklabels={0,,250k,,500k},
      yticklabels=\empty,
      ymin=0,ymax=100,
      ymajorgrids=true,
      cycle list name=custom cycle list,
    ]
    \addplot table [x=episode,y=hit_rate,col sep=comma] {data/symmetry/o3_sym0.txt};
    \addplot table [x=episode,y=hit_rate,col sep=comma] {data/symmetry/o3_sym1.txt};
  \end{axis}
\end{tikzpicture}%

  \caption{Impact of symmetry enforcement on performance metrics during training.}
  \label{fig:ablation_symmetry}
\end{figure}
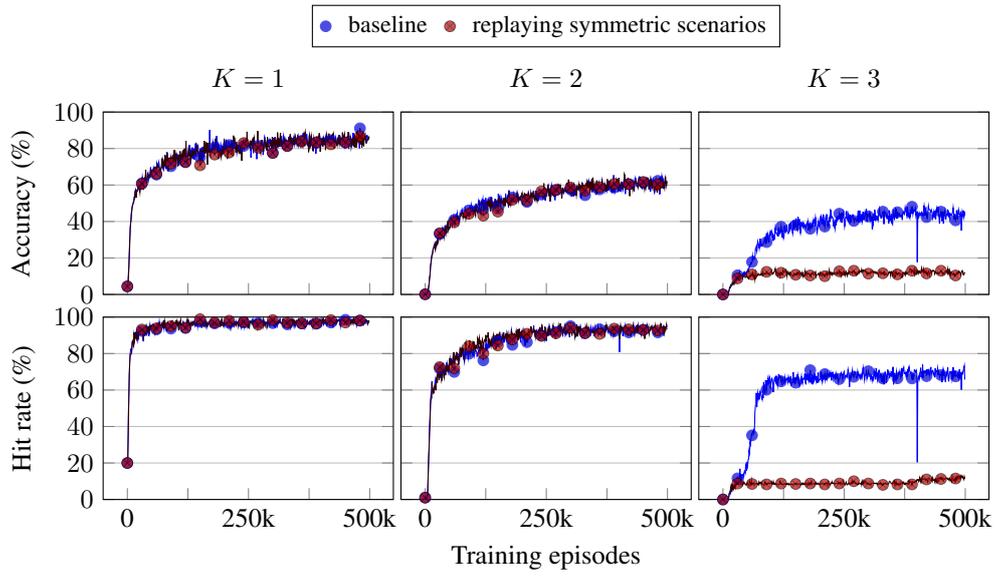

Our framework leverages the reciprocity principle of electromagnetic wave propagation, which states that paths from a transmitter to a receiver are equivalent to paths from a receiver to a transmitter. We hypothesize that by enforcing this symmetry during training---doubling each sampled path by including both the forward and reverse directions---the model can learn more efficiently from fewer unique geometric configurations. We evaluate whether enforcing this symmetry accelerates convergence and improves generalization compared to standard training, which treats each direction independently. \Cref{fig:ablation_symmetry} shows the training curves for these configurations and illustrates how each component contributes to overall performance.

\paragraph{Importance of the Training Scenario}

The diversity and realism of training scenarios directly influence the ability of the model to generalize to unseen configurations. We investigate the impact of two key training scenario variations: (1) the spatial region from which transmitter and receiver positions are sampled (canyon-only versus whole scene), and (2) the randomization of scene geometry through building removal.

In our previous work~\cite{icmlcn2025}, we generated training variations by randomly removing individual triangular facets from buildings. While this approach provided geometric diversity, we observed that it produced unrealistic scenes---for example, with large chunks missing from building facades---which appeared to negatively impact model performance. To address this limitation, we now adopt a per-building removal strategy: instead of removing facets, we randomly exclude entire buildings from each training sample. This approach maintains geometric realism while still providing sufficient scene diversity to prevent overfitting to specific architectures.

\begin{figure}
  \centering
  \tikzsetnextfilename{tikzexternalize/ablation_sampling_region}%
  \begin{tikzpicture}[every plot/.append style={mark repeat=30},every mark/.append style={opacity=0.7}]
  \newcommand{\plotwidth}{5.4cm}
  \newcommand{\plotheight}{4cm}
  \pgfplotscreateplotcyclelist{custom cycle list}{
    blue,every mark/.append style={fill=blue!80!black},mark=*\\
    red!20!black,every mark/.append style={fill=red!80!black},mark=otimes*\\
    brown,every mark/.append style={fill=brown!80!black},mark=square*\\
    gray,every mark/.append style={fill=gray!80!black},mark=diamond*\\
  }
  \pgfplotsset{every axis/.append style={scaled x ticks=false}}
  \pgfplotsset{every x tick label/.append style={/pgf/number format/fixed}}

  % Accuracy

  \begin{axis}[
      name=plot_order_1_sr,
      width=\plotwidth,
      height=\plotheight,
      ylabel={Accuracy (\%)},
      title={$K=1$},
      xtick pos=left,
      xtick={0,1.25e5, 2.5e5, 3.75e5, 5e5},
      xticklabels=\empty,
      ymin=0,ymax=100,
      ymajorgrids=true,
      cycle list name=custom cycle list,
    ]
    \addplot table [x=episode,y=success_rate,col sep=comma] {data/sampling_region/o1_sc1_f1.txt};
    \addplot table [x=episode,y=success_rate,col sep=comma] {data/sampling_region/o1_sc1_f0.txt};
    \addplot table [x=episode,y=success_rate,col sep=comma] {data/sampling_region/o1_sc0_f1.txt};
    \addplot table [x=episode,y=success_rate,col sep=comma] {data/sampling_region/o1_sc0_f0.txt};
  \end{axis}

  \begin{axis}[
      name=plot_order_2_sr,
      at={($ (.1cm,0cm) + (plot_order_1_sr.south east) $)},
      width=\plotwidth,
      height=\plotheight,
      title={$K=2$},
      title style={name=title},
      xtick pos=left,
      xtick={0,1.25e5, 2.5e5, 3.75e5, 5e5},
      xticklabels=\empty,
      yticklabels=\empty,
      ymin=0,ymax=100,
      ymajorgrids=true,
      cycle list name=custom cycle list,
      legend style={font=\small, at={([yshift=1ex]title.north)},anchor=south,name=leg,nodes={anchor=base}},
      legend columns=-1,legend style={column sep=1ex},
      legend image post style={yshift=.5ex},
      every legend to name picture/.style={
        baseline={(leg.base)},
      }
    ]
    \addlegendimage{blue,only marks,fill=blue!80!black,mark=*}
    \addlegendimage{red!20!black,only marks,fill=red!80!black,mark=otimes*}
    \addlegendimage{brown,only marks,fill=brown!80,mark=square*}
    \addlegendimage{gray,only marks,fill=gray!80!black,mark=diamond*}
    \legend{canyon-only (CO),CO + optional ground (OG),whole scene (WS),WS+OG}
    \addplot table [x=episode,y=success_rate,col sep=comma] {data/sampling_region/o2_sc1_f1.txt};
    \addplot table [x=episode,y=success_rate,col sep=comma] {data/sampling_region/o2_sc1_f0.txt};
    \addplot table [x=episode,y=success_rate,col sep=comma] {data/sampling_region/o2_sc0_f1.txt};
    \addplot table [x=episode,y=success_rate,col sep=comma] {data/sampling_region/o2_sc0_f0.txt};
  \end{axis}

  \begin{axis}[
      name=plot_order_3_sr,
      at={($ (.1cm,0cm) + (plot_order_2_sr.south east) $)},
      width=\plotwidth,
      height=\plotheight,
      title={$K=3$},
      xtick pos=left,
      xtick={0,1.25e5, 2.5e5, 3.75e5, 5e5},
      xticklabels=\empty,
      yticklabels=\empty,
      ymin=0,ymax=100,
      ymajorgrids=true,
      cycle list name=custom cycle list,
    ]
    \addplot table [x=episode,y=success_rate,col sep=comma] {data/sampling_region/o3_sc1_f1.txt};
    \addplot table [x=episode,y=success_rate,col sep=comma] {data/sampling_region/o3_sc1_f0.txt};
    \addplot table [x=episode,y=success_rate,col sep=comma] {data/sampling_region/o3_sc0_f1.txt};
    \addplot table [x=episode,y=success_rate,col sep=comma] {data/sampling_region/o3_sc0_f0.txt};
  \end{axis}

  % Hit rate

  \begin{axis}[
      name=plot_order_1_hr,
      anchor=north,
      at={($ (0cm,-.3cm) + (plot_order_1_sr.south) $)},
      width=\plotwidth,
      height=\plotheight,
      ylabel={Hit rate (\%)},
      xtick pos=left,
      xtick={0,1.25e5, 2.5e5, 3.75e5, 5e5},
      xticklabels={0,,250k,,500k},
      x tick label style={/pgf/number format/fixed},
      ymin=0,ymax=100,
      ymajorgrids=true,
      cycle list name=custom cycle list,
    ]
    \addplot table [x=episode,y=hit_rate,col sep=comma] {data/sampling_region/o1_sc1_f1.txt};
    \addplot table [x=episode,y=hit_rate,col sep=comma] {data/sampling_region/o1_sc1_f0.txt};
    \addplot table [x=episode,y=hit_rate,col sep=comma] {data/sampling_region/o1_sc0_f1.txt};
    \addplot table [x=episode,y=hit_rate,col sep=comma] {data/sampling_region/o1_sc0_f0.txt};
  \end{axis}

  \begin{axis}[
      name=plot_order_2_hr,
      at={($ (.1cm,0cm) + (plot_order_1_hr.south east) $)},
      width=\plotwidth,
      height=\plotheight,
      xlabel={Training episodes},
      xtick pos=left,
      xtick={0,1.25e5, 2.5e5, 3.75e5, 5e5},
      xticklabels={0,,250k,,500k},
      yticklabels=\empty,
      ymin=0,ymax=100,
      ymajorgrids=true,
      cycle list name=custom cycle list,
    ]
    \addplot table [x=episode,y=hit_rate,col sep=comma] {data/sampling_region/o2_sc1_f1.txt};
    \addplot table [x=episode,y=hit_rate,col sep=comma] {data/sampling_region/o2_sc1_f0.txt};
    \addplot table [x=episode,y=hit_rate,col sep=comma] {data/sampling_region/o2_sc0_f1.txt};
    \addplot table [x=episode,y=hit_rate,col sep=comma] {data/sampling_region/o2_sc0_f0.txt};
  \end{axis}

  \begin{axis}[
      name=plot_order_3_hr,
      at={($ (.1cm,0cm) + (plot_order_2_hr.south east) $)},
      width=\plotwidth,
      height=\plotheight,
      xtick pos=left,
      xtick={0,1.25e5, 2.5e5, 3.75e5, 5e5},
      xticklabels={0,,250k,,500k},
      yticklabels=\empty,
      ymin=0,ymax=100,
      ymajorgrids=true,
      cycle list name=custom cycle list,
    ]
    \addplot table [x=episode,y=hit_rate,col sep=comma] {data/sampling_region/o3_sc1_f1.txt};
    \addplot table [x=episode,y=hit_rate,col sep=comma] {data/sampling_region/o3_sc1_f0.txt};
    \addplot table [x=episode,y=hit_rate,col sep=comma] {data/sampling_region/o3_sc0_f1.txt};
    \addplot table [x=episode,y=hit_rate,col sep=comma] {data/sampling_region/o3_sc0_f0.txt};
  \end{axis}
\end{tikzpicture}%

  \caption{Training scenario impact on performance metrics during training.}
  \label{fig:ablation_sampling_region}
\end{figure}
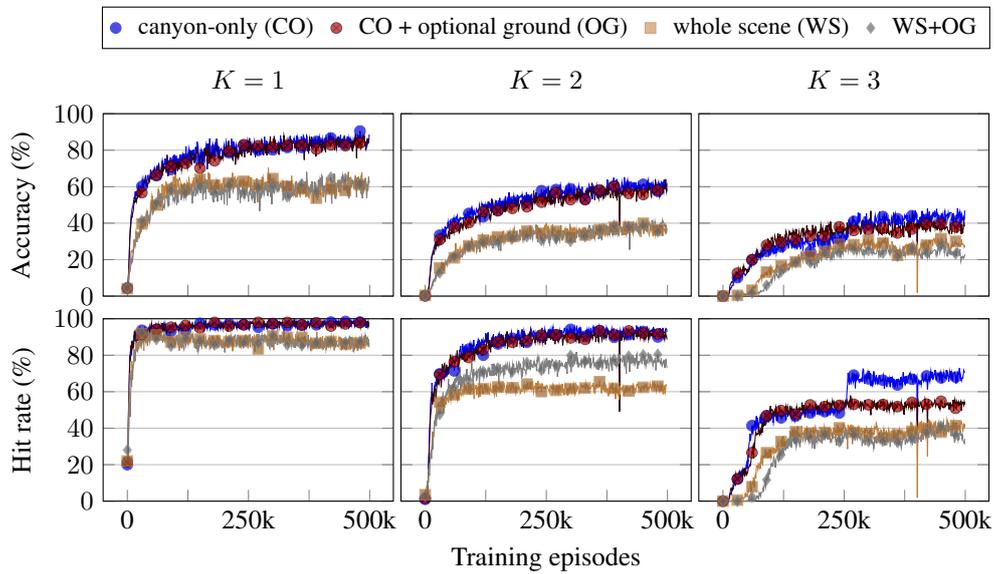

By comparing models trained under these different scenario generation strategies, we assess the trade-off between training on highly constrained, canonical configurations versus more diverse but potentially noisier data distributions. \Cref{fig:ablation_sampling_region} presents how each component contributes to overall performance for the improved per-building removal approach used throughout this work. The previous facet-removal method can be enabled in the provided code for comparison.

\subsubsection{Simulation Time}\label{sec:benchmarks}

\begin{figure}
  \centering
  \tikzsetnextfilename{tikzexternalize/benchmarks}%
  \begin{tikzpicture}[every plot/.append style={mark repeat=5},every mark/.append style={opacity=0.7}]
  \newcommand{\plotwidth}{5.4cm}
  \newcommand{\plotheight}{5cm}
  \newcommand{\xmode}{log} % linear or log
  \pgfplotscreateplotcyclelist{custom cycle list}{
    blue,every mark/.append style={fill=blue!80!black},mark=*\\
    red!20!black,every mark/.append style={fill=red!80!black},mark=otimes*\\
    brown,every mark/.append style={fill=brown!80!black},mark=square*\\
    gray,every mark/.append style={fill=gray!80!black},mark=diamond*\\
    mark=star\\
  }
  \pgfplotsset{every axis/.append style={scaled x ticks=false}}
  \pgfplotsset{every x tick label/.append style={/pgf/number format/fixed}}

  % CPU time

  \begin{axis}[
      name=plot_order_1_cpu,
      width=\plotwidth,
      height=\plotheight,
      ylabel={CPU time (\unit{\milli\second})},
      title={$K=1$},
      xtick pos=left,
      xticklabels=\empty,
      ymin=0.5e-1,ymax=2e3,
      ymajorgrids=true,
      xmode=\xmode,
      ymode=log,
      cycle list name=custom cycle list,
    ]
    \pgfplotstableread[col sep=comma]{data/benchmarks/cpu_benchmarks_1.txt}\mydata;
    \addplot table [x=n,y expr=\thisrow{bs_1} * 1000] {\mydata};
    \addplot table [x=n,y expr=\thisrow{bs_10} * 1000] {\mydata};
    \addplot table [x=n,y expr=\thisrow{bs_100} * 1000] {\mydata};
    \addplot table [x=n,y expr=\thisrow{bs_1000} * 1000] {\mydata};
    \addplot table [x=n,y expr=\thisrow{exhaustive} * 1000] {\mydata};
  \end{axis}

  \begin{axis}[
      name=plot_order_2_cpu,
      at={($ (.1cm,0cm) + (plot_order_1_cpu.south east) $)},
      width=\plotwidth,
      height=\plotheight,
      title={$K=2$},
      title style={name=title},
      xtick pos=left,
      xticklabels=\empty,
      yticklabels=\empty,
      ymin=0.5e-1,ymax=2e3,
      ymajorgrids=true,
      xmode=\xmode,
      ymode=log,
      cycle list name=custom cycle list,
      legend style={font=\small, at={([yshift=1ex]title.north)},anchor=south,name=leg,nodes={anchor=base}},
      legend columns=-1,legend style={column sep=1ex},
      legend image post style={yshift=.5ex},
      every legend to name picture/.style={
        baseline={(leg.base)},
      }
    ]
    \addlegendimage{blue,only marks,fill=blue!80!black,mark=*}
    \addlegendimage{red!20!black,only marks,fill=red!80!black,mark=otimes*}
    \addlegendimage{brown,only marks,fill=brown!80,mark=square*}
    \addlegendimage{gray,only marks,fill=gray!80!black,mark=diamond*}
    \addlegendimage{only marks,fill=black,mark=star}
    \legend{$M=1$,$M=10$,$M=100$,$M=1000$,exhaustive}
    \pgfplotstableread[col sep=comma]{data/benchmarks/cpu_benchmarks_2.txt}\mydata;
    \addplot table [x=n,y expr=\thisrow{bs_1} * 1000] {\mydata};
    \addplot table [x=n,y expr=\thisrow{bs_10} * 1000] {\mydata};
    \addplot table [x=n,y expr=\thisrow{bs_100} * 1000] {\mydata};
    \addplot table [x=n,y expr=\thisrow{bs_1000} * 1000] {\mydata};
    \addplot table [x=n,y expr=\thisrow{exhaustive} * 1000] {\mydata};
  \end{axis}

  \begin{axis}[
      name=plot_order_3_cpu,
      at={($ (.1cm,0cm) + (plot_order_2_cpu.south east) $)},
      width=\plotwidth,
      height=\plotheight,
      title={$K=3$},
      xtick pos=left,
      xticklabels=\empty,
      yticklabels=\empty,
      ymin=0.5e-1,ymax=2e3,
      ymajorgrids=true,
      xmode=\xmode,
      ymode=log,
      cycle list name=custom cycle list,
      set layers,
    ]
    \pgfplotstableread[col sep=comma]{data/benchmarks/cpu_benchmarks_3.txt}\mydata;
    \addplot table [x=n,y expr=\thisrow{bs_1} * 1000] {\mydata};
    \addplot table [x=n,y expr=\thisrow{bs_10} * 1000] {\mydata};
    \addplot table [x=n,y expr=\thisrow{bs_100} * 1000] {\mydata};
    \addplot table [x=n,y expr=\thisrow{bs_1000} * 1000] {\mydata};
    \addplot table [x=n,y expr=\thisrow{exhaustive} * 1000] {\mydata};
    \begin{pgfonlayer}{axis foreground}
      \draw[|<->|,dashed] (axis cs:9e1,1e1) -- (axis cs:9e1,1e3);
      \node[anchor=north west,font=\scriptsize,fill=white,fill opacity=0.85,text opacity=1,inner sep=1pt] at (rel axis cs:0.03,0.97) {\shortstack{$\approx 10^2\times$ faster\\at $M=10$}};
    \end{pgfonlayer}
  \end{axis}

  % GPU time

  \begin{axis}[
      name=plot_order_1_gpu,
      anchor=north,
      at={($ (0cm,-.3cm) + (plot_order_1_cpu.south) $)},
      width=\plotwidth,
      height=\plotheight,
      ylabel={GPU time (\unit{\milli\second})},
      xtick pos=left,
      x tick label style={/pgf/number format/fixed},
      ymin=1.5e-1,ymax=2e2,
      ymajorgrids=true,
      xmode=\xmode,
      ymode=log,
      cycle list name=custom cycle list,
    ]
    \pgfplotstableread[col sep=comma]{data/benchmarks/gpu_benchmarks_1.txt}\mydata;
    \addplot table [x=n,y expr=\thisrow{bs_1} * 1000] {\mydata};
    \addplot table [x=n,y expr=\thisrow{bs_10} * 1000] {\mydata};
    \addplot table [x=n,y expr=\thisrow{bs_100} * 1000] {\mydata};
    \addplot table [x=n,y expr=\thisrow{bs_1000} * 1000] {\mydata};
    \addplot table [x=n,y expr=\thisrow{exhaustive} * 1000] {\mydata};
  \end{axis}

  \begin{axis}[
      name=plot_order_2_gpu,
      at={($ (.1cm,0cm) + (plot_order_1_gpu.south east) $)},
      width=\plotwidth,
      height=\plotheight,
      xlabel={Number of scene objects},
      xtick pos=left,
      yticklabels=\empty,
      ymin=1.5e-1,ymax=2e2,
      ymajorgrids=true,
      xmode=\xmode,
      ymode=log,
      cycle list name=custom cycle list,
    ]
    \pgfplotstableread[col sep=comma]{data/benchmarks/gpu_benchmarks_2.txt}\mydata;
    \addplot table [x=n,y expr=\thisrow{bs_1} * 1000] {\mydata};
    \addplot table [x=n,y expr=\thisrow{bs_10} * 1000] {\mydata};
    \addplot table [x=n,y expr=\thisrow{bs_100} * 1000] {\mydata};
    \addplot table [x=n,y expr=\thisrow{bs_1000} * 1000] {\mydata};
    \addplot table [x=n,y expr=\thisrow{exhaustive} * 1000] {\mydata};
  \end{axis}

  \begin{axis}[
      name=plot_order_3_gpu,
      at={($ (.1cm,0cm) + (plot_order_2_gpu.south east) $)},
      width=\plotwidth,
      height=\plotheight,
      xtick pos=left,
      yticklabels=\empty,
      ymin=1.5e-1,ymax=2e2,
      ymajorgrids=true,
      xmode=\xmode,
      ymode=log,
      cycle list name=custom cycle list,
      set layers,
    ]
    \pgfplotstableread[col sep=comma]{data/benchmarks/gpu_benchmarks_3.txt}\mydata;
    \addplot table [x=n,y expr=\thisrow{bs_1} * 1000] {\mydata};
    \addplot table [x=n,y expr=\thisrow{bs_10} * 1000] {\mydata};
    \addplot table [x=n,y expr=\thisrow{bs_100} * 1000] {\mydata};
    \addplot table [x=n,y expr=\thisrow{bs_1000} * 1000] {\mydata};
    \addplot table [x=n,y expr=\thisrow{exhaustive} * 1000] {\mydata};
    \begin{pgfonlayer}{axis foreground}
      \draw[|<->|,dashed] (axis cs:9e1,2e0) -- (axis cs:9e1,4e1);
      \node[anchor=north west,font=\scriptsize,fill=white,fill opacity=0.85,text opacity=1,inner sep=1pt] at (rel axis cs:0.03,0.97) {\shortstack{$\approx 10^1\times$ faster\\at $M=10$}};
    \end{pgfonlayer}
  \end{axis}
\end{tikzpicture}%

  \caption{Benchmarks of the computation time required to perform a full ray tracing simulation as a function of the number of objects in the scene for interaction orders $K \in \{1, 2, 3\}$. Each plot compares the CPU time (top row) and GPU time (bottom row) of our machine-learning-assisted approach, for various sampling budgets $M$, against an exhaustive ray tracing baseline. For $M=10$, our method reaches speedups of roughly two orders of magnitude on CPU and one order of magnitude on GPU over exhaustive search.}
  \label{fig:benchmarks}
\end{figure}
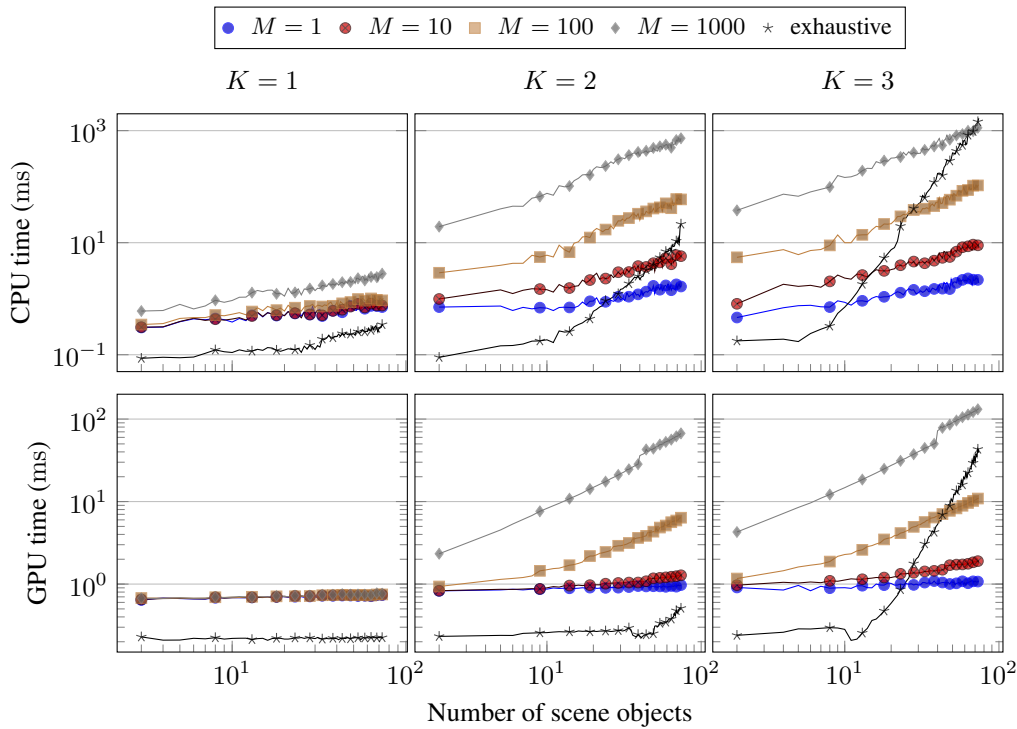

Computational efficiency represents the primary motivation for our approach. While achieving high hit rates demonstrates that our model successfully learns to identify valid paths, the practical value of the framework depends critically on whether this learned sampling strategy translates into meaningful speedups over exhaustive enumeration. As discussed in \cref{sec:concept}, the theoretical advantage of our method depends on whether a CPU or GPU is used, the number of objects in the scene, and the interaction order $K$. Here, we empirically validate these claims by measuring the execution time on differently sized scenes and comparing against a highly optimized baseline.

We compare the runtime of our machine-learning-assisted approach against a highly optimized exhaustive ray tracing baseline implemented in JAX (part of the DiffeRT library). For a fair comparison, both methods utilize just-in-time compilation and are executed on the same hardware configurations: Intel\textcopyright{} Xeon\textcopyright{} W-1370P $\times$ 16 (CPU) with a NVIDIA GeForce RTX\texttrademark{} 3070 (GPU). We do not compare against \gls{sbr} approaches here because they usually target fast, but approximate coverage-map generation rather than exhaustive point-to-point path enumeration, which is the problem addressed in this paper.

The results are shown in \cref{fig:benchmarks} for interaction orders $K$ ranging from 1 to 3. Each plot displays the total computation time required to perform a full ray tracing simulation as a function of the number of objects in the scene. The machine-learning-assisted approach is evaluated for various sampling budgets $M$, while the exhaustive baseline is shown for reference.

\subsubsection{Coverage Map Prediction}

While hit rates and computational efficiency are useful intermediate metrics, the key test is whether the method can predict accurate radio coverage maps for practical planning. Unlike isolated path sampling, coverage prediction requires a representative set of paths whose combined contribution approximates the total received field. We therefore compare predicted coverage maps against \gls{gt} maps generated by exhaustive ray tracing. Specifically, we compute two maps on a receiver grid at \qty{1.5}{\meter} above ground: the \gls{gt} map from exhaustive search and the predicted map obtained by sampling path candidates with our model. The model is trained for \mbox{first-}, \mbox{second-}, and third-order specular reflections. For the line-of-sight component, no model is used, since at most one line-of-sight path exists between two antennas regardless of geometry.

To compare maps, we compute the total path gain in decibels (\unit{\dB}), i.e., the negative of path loss, and define two metrics: the relative residual map, defined as
\begin{equation}\label{eq:absd}
  \Delta G_{\mathrm{r}} = \frac{G_\mathrm{\glsxtrshort{gt}}-G_\mathrm{pred}}{G_\mathrm{\glsxtrshort{gt}}},
\end{equation}
and the relative residual map in decibels, given by
\begin{equation}\label{eq:reld}
  \Delta G_{\mathrm{r},\unit{\dB}} = 10\log_{10}\left(\frac{G_\mathrm{\glsxtrshort{gt}}}{G_\mathrm{pred}}\right).
\end{equation}

\paragraph{Idealized Street-Canyon Scenario}

\begin{figure}[htbp]
  \centering
  \subfloat[Ground truth $G_\mathrm{\glsxtrshort{gt}}$ (\unit{\dB}).]{\label{fig:gt}\import{pgf}{exhau.pgf}}
  \subfloat[Prediction $G_\mathrm{pred}$ (\unit{\dB}).]{\label{fig:pred}\import{pgf}{pred.pgf}}
  \\
  \subfloat[Relative difference $\Delta G_{\mathrm{r}}$.]{\label{fig:absd}\import{pgf}{diff.pgf}}
  \subfloat[Log-relative difference $\Delta G_{\mathrm{r},\unit{\dB}}$.]{\label{fig:reld}\import{pgf}{rdiff.pgf}}
  \caption{Comparison of the coverage map between the ground truth (\protect\cref{fig:gt}) and the prediction (\protect\cref{fig:pred}) using a representative budget of $M=20$. \protect\Cref{fig:absd,fig:reld} show the relative and log-relative differences (in \unit{\dB}) between the two coverage maps, as defined in \cref{eq:absd,eq:reld}.}
  \label{fig:coverage_map}
\end{figure}

\Cref{fig:coverage_map} shows the ground truth and predicted coverage maps, as well as the absolute and relative residuals, for a transmitting antenna located at $\boldsymbol{x}_{\mathrm{\glsxtrshort{tx}}} =
\begin{bmatrix} \qty{0}{\meter} & \qty{0}{\meter} & \qty{32}{\meter}
\end{bmatrix}^\top$. To avoid frequency-dependent effects, the path gain is set to the inverse squared path length and reflection coefficients are set to 1 for all paths, emphasizing the impact of sampling reflection paths correctly.

\Cref{fig:absd,fig:reld} are complementary. The relative difference shows the relative contribution of valid paths found by the model and highlights the locations where the model does not find all possible ray paths using a limited number of samples ($M = 20$), while the log-relative difference provides a more intuitive visualization of the relative error in decibels. The absence of color in \cref{fig:reld} does not necessarily mean that the model found all possible solutions; it can also indicate that the model did not find any valid path, which leads to a division by zero in \eqref{eq:absd}. In theory, the choice of the sampling budget depends on desired performance and computational constraints, but also on the number of objects in the scene. Here, we observe that the hit rate would increase with an increased sampling budget, but rapidly saturate after $M=10$ in most cases. We therefore chose $M=20$ as a balance between performance and computational efficiency.

\paragraph{Generalization to a Realistic Manhattan Scenario}

\begin{figure}
  \centering
  \subfloat[Small scene ($N=400$).]{\label{fig:manhattan_small}\includegraphics[width=.45\textwidth]{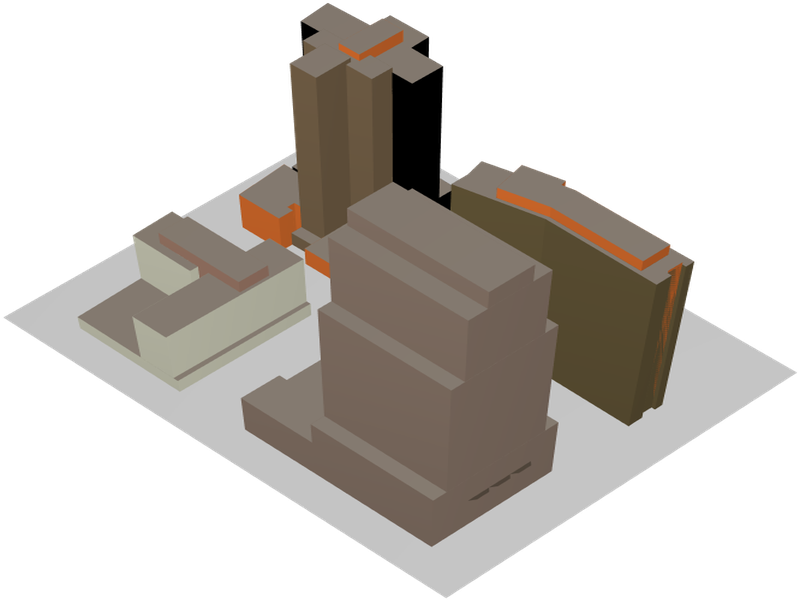}}
  \subfloat[Medium scene ($N=805$).]{\label{fig:manhattan_medium}\includegraphics[width=.45\textwidth]{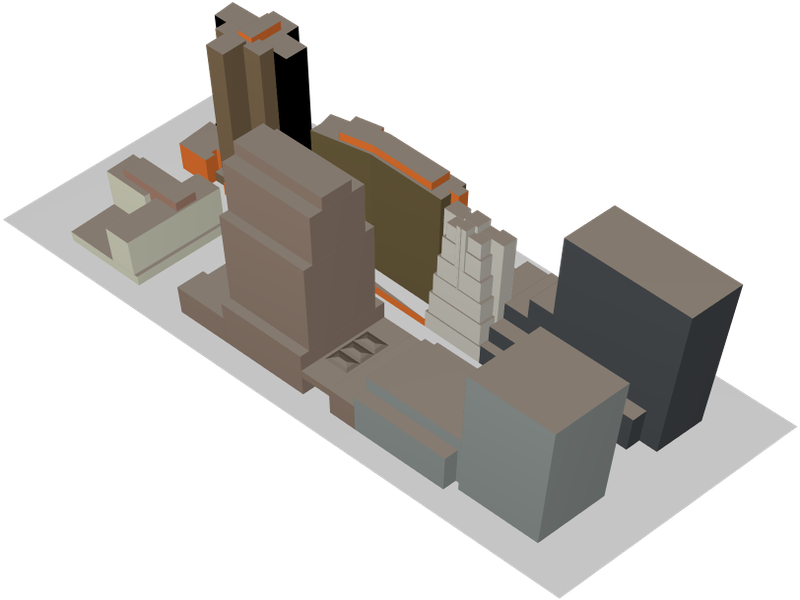}}
  \caption{Street canyon observed in Manhattan, obtained from OpenStreetMap.}
  \label{fig:manhattan_scenes}
\end{figure}

While the results in \cref{fig:coverage_map} demonstrate that our model can learn to sample valid paths and predict coverage maps in the idealized street canyon scenario, an important limitation of this evaluation must be acknowledged: both the training and validation sets were generated as procedural variations of a single canonical street canyon geometry. Although this generation process introduces diversity through random building removal, variable transmitter-receiver placements, and optional ground plane inclusion, the resulting scenes remain fundamentally drawn from the same distributional family. This raises the question of whether the model has learned generalizable principles of path geometry, or whether it has primarily adapted to the specific structural characteristics of this particular canyon topology.

To address this concern, we evaluate the trained model on a substantially different urban environment: real-world street canyon scenes obtained from OpenStreetMap data in Manhattan, New York. These scenes exhibit structural characteristics distinct from the training distribution, including different building aspect ratios, facade complexity, street widths, and block organization patterns. While still maintaining the overall street-canyon form factor, these real-world geometries provide a genuine out-of-distribution test case.

The Manhattan scenes include a larger number of objects ($N=400$ and $N=805$ for the small and medium scenes, respectively) compared to the idealized scenario (up to $N=74$), which dramatically increases the number of possible paths. Additionally, the real-world scenes exhibit finer and more complex building details not present in the idealized procedurally-generated scenario. These factors substantially increase geometric complexity and multipath density. The scenes are obtained from OpenStreetMap and are shown in \cref{fig:manhattan_scenes}. Due to the increased complexity of the scene, only up to second-order reflections are considered in this evaluation to maintain a reasonable computational cost for the exhaustive search (which is used to compute the ground truth). The sampling budget is also set to $M=100$ to account for the increased number of valid paths in this more complex scenario. The results are shown in \cref{fig:coverage_map_manhattan}.

\begin{figure}[htbp]
  \centering
  \subfloat[Ground truth $G_\mathrm{\glsxtrshort{gt}}$ (\unit{\dB}) (small scene).]{\label{fig:gt-manhattan-small}\import{pgf}{manhattan_small_exhau.pgf}}
  \subfloat[Prediction $G_\mathrm{pred}$ (\unit{\dB}) (small scene).]{\label{fig:pred-manhattan-small}\import{pgf}{manhattan_small_pred.pgf}}
  \\
  \subfloat[Ground truth $G_\mathrm{\glsxtrshort{gt}}$ (\unit{\dB}) (medium scene).]{\label{fig:gt-manhattan-medium}\import{pgf}{manhattan_medium_exhau.pgf}}
  \subfloat[Prediction $G_\mathrm{pred}$ (\unit{\dB}) (medium scene).]{\label{fig:pred-manhattan-medium}\import{pgf}{manhattan_medium_pred.pgf}}
  \caption{Comparison of the coverage map between the ground truth (\protect\cref{fig:gt}) and the prediction (\protect\cref{fig:pred}) using a sampling budget of $M=100$.}
  \label{fig:coverage_map_manhattan}
\end{figure}

\paragraph{Practical note on coverage-map generation}

It is important to emphasize that exhaustive point-to-point ray tracing is not an efficient approach for producing coverage maps in practical deployments: exhaustive enumeration computes paths independently for each receiver location and therefore scales poorly when many receiver positions must be evaluated. \Gls{sbr} and other ray-launching techniques are typically far more efficient for coverage-map generation because the same launched rays can be reused across multiple receiver positions. Our learned sampler targets the complementary problem of accelerating exhaustive point-to-point enumeration (for tasks that require full path lists or differentiable path computations) rather than replacing \gls{sbr}-style methods for bulk coverage-map production. Accordingly, coverage maps in this work are used for visualization and for evaluating sampler fidelity against an exhaustive ground truth, not as a recommendation to use exhaustive tracing for routine coverage-map generation.

\subsection{Discussion}

Our experimental evaluation of the machine-learning-assisted ray tracing framework in the urban street canyon application demonstrates that the proposed approach successfully achieves the dual objectives of high sampling accuracy and computational efficiency. In this section, we synthesize the key findings, discuss their implications for practical deployment, acknowledge important limitations, and outline directions for future work.

\subsubsection{Effectiveness of the Proposed Framework}\label{sec:framework_effectiveness}

The ablation study reveals that each component of our framework contributes positively to overall performance, but with varying degrees of impact. As illustrated in \cref{fig:ablation_buffer_epsilon}, the addition of the replay buffer is crucial for allowing the model to converge to a relatively high hit rate, even in the early stages of training and for higher interaction orders. The exploratory policy has a more moderate impact on the training performance. We observe that using $\epsilon$-greedy exploration enables the model to explore more solutions (leading to higher hit rates), at the potential cost of a lower accuracy (potentially linked to overfitting). \Cref{fig:ablation_masking_weighting} reveals that action masking has a very limited impact on the training performance, by only slightly improving the performance results for $K=3$. This could be explained by the fact that the visibility-based technique behind the action masking strategy is relatively simple. Distance-based flow weighting has a negative impact, particularly for lower interaction orders, where it decreases convergence speed and performance results even more. Furthermore, we observe that the model collapses in most cases after a number of training episodes. This suggests that biasing the sampling distribution toward geometrically shorter paths may not always be an effective heuristic, likely because it is not directly linked to the output of the reward function: a valid ray path is not necessarily short. \Cref{fig:ablation_symmetry} shows that enforcing symmetry has a very negligible impact on the training performance for $K=1$ and $K=2$, and has a severe negative impact for $K=2$. This could be explained by the fact that (1) the symmetry is not promoted anywhere in the reward function, meaning that the model does not receive any additional signal if it provides those important symmetric properties, and (2) replaying symmetric experiences also indirectly reduces the diversity of the scenes on which the model is trained. Finally, \cref{fig:ablation_sampling_region} shows that training exclusively on the street canyon leads to better generalization within the tested street-canyon family and higher hit rates compared to training on the whole scene, likely due to the reduced diversity of geometric configurations encountered during training.

Except for the replay buffer, which is instrumental to the convergence, the other components provide varying degrees of improvement, with some even showing trade-offs in certain scenarios. This highlights the importance of carefully designing and tuning each component to achieve optimal performance.

Finally, the training curves demonstrate that for $K=1$ and $K=2$, the model rapidly reaches a hit rate above \qty{90}{\percent}. This represents a substantial improvement over our previous work~\cite{icmlcn2025}---primarily due to solving the sparse-reward problem and preventing model collapse. For $K=3$, the model achieves almost \qty{45}{\percent} accuracy and a \qty{65}{\percent} hit rate, successfully overcoming previously faced problems as well. However, the fact that the hit rate for $K=3$ converges below \qty{90}{\percent} suggests that the model does not fully learn to generalize across all possible configurations. More precisely, the current evidence supports reasonable generalization within the randomized street-canyon family used in this paper, but not yet across arbitrary urban morphologies or indoor scenes. This limitation is illustrated in \cref{fig:coverage_map}, where we observe that some higher-order reflection regions are entirely missing. An important consideration when interpreting these results is that hit rates are evaluated using only $M=10$ samples per scene, which is particularly constraining for $K=3$ where valid paths are extremely rare (see \cref{tab:scene_statistics}). We acknowledge that this small sampling budget might underestimate the true coverage capability of the model, and increasing $M$ may reveal better performance than currently observed. To bridge the remaining performance gap for higher interaction orders, curriculum learning (e.g., training on $K=1$ first and then increasing to $K=3$) or stratified/adaptive sampling could be explored, and further hyperparameter fine-tuning may also be beneficial. Moreover, these observations motivate future work on exploring more expressive architectures, such as transformer-based models or graph neural networks, which could better capture the complex relational information between objects.

\subsubsection{Quality of Coverage Prediction}

The results in the idealized street-canyon scenario indicate that our model approximates the exhaustive search reasonably well---yielding an RMSE of \qty{3.34}{\dB} for the full scene and \qty{1.51}{\dB} for the main canyon (see \cref{fig:reld})---despite generating only a fraction of possible path candidates. However, visual inspection reveals that while the model captures clear reflection patterns, it fails to reconstruct entire second- and especially third-order reflection regions. The lack of apparent noise suggests the model overfits to dominant paths and neglects the exploration of alternative geometric solutions. Additionally, the residual error observed in the immediate vicinity of the transmitter can be attributed to the geometric transformation, which approaches a singularity when the \gls{rx} shares the same horizontal coordinates as the \gls{tx}. While we explicitly avoid division by zero, numerical instabilities in this region persist.

When evaluated on the out-of-distribution Manhattan scenes, the generalization limitations become more pronounced. While the model continues to identify certain dominant reflection paths, the coverage maps in the Manhattan scenarios exhibit significant spatial variation in prediction quality. Some regions are well-captured, but others---including areas with first-order reflections---show substantial discrepancies from the ground truth. Notably, the model completely fails to discover valid reflection paths from certain building facades, and for other facades, valid paths are only recovered in localized spatial regions. This spatial heterogeneity suggests that the model has not learned robust, generalizable principles for identifying reflections, but rather has overfitted to the specific geometric configurations and reflection patterns present in the idealized training scenarios. These results underscore that additional work is necessary before the current model can be reliably deployed on real-world scenes with arbitrary geometries.

\subsubsection{Implications for Practical Deployment}

The achieved hit rate exceeding \qty{90}{\percent} represents a transformative improvement over the $\ll \qty{1}{\percent}$ hit rates from naive random sampling. In practical terms, this means that computational resources are deployed far more efficiently: rather than exhaustively enumerating hundreds of thousands of candidate paths to find a representative sample, the model identifies valid paths using a relatively small sampling budget. This efficiency gain translates directly to reduced runtime in production systems. For training, however, the upfront cost remains non-negligible: with \num{500000} iterations, a single model typically requires about 45 minutes to 1.5 hours on an RTX\texttrademark{} 3070 GPU. Our method should therefore be understood as an amortized accelerator: it is most beneficial when the same scene family is queried repeatedly or when many coverage evaluations are needed. We do not claim zero-shot transfer to arbitrary unseen scenes.

However, realizing these benefits requires sufficiently similar deployment scenarios. A critical practical challenge is the need for procedurally generated synthetic training data: creating diverse, realistic training scenarios for new deployment environments is non-trivial and often requires domain-specific expertise. For scenarios substantially different from our urban street canyon training class---such as different urban morphologies, indoor environments, or propagation phenomena beyond specular reflections---one must either retrain the model on newly generated data or resort to exhaustive ray tracing. In such cases, the training cost may exceed the computational savings from a single or a few ray tracing simulations, making exhaustive enumeration more practical.

\subsubsection{Computational Advantages and Limitations}

Our benchmark results confirm the theoretical predictions from \cref{sec:concept}: speedups are less pronounced on GPU-accelerated hardware, where the parallelism usually mitigates the need to explore many path candidates. On CPUs, where this parallelism is limited, the ability to reduce the number of path candidates is particularly valuable. However, as the number of scene objects and interaction order increase, we see that our model becomes actually faster than an exhaustive search, even on relatively small scenes such as the one studied here.

As scene complexity and interaction order grow, the combinatorial cost of exhaustive enumeration increases rapidly, so our learned sampler can become comparatively faster even on GPU-equipped hardware for moderately sized scenes.

A critical observation is that the magnitude of speedups depends heavily on the acceptance ratio of the baseline method. In scenes where geometrically valid paths are rare---as in our urban street canyon with its constrained multipath propagation---intelligent sampling provides substantial benefits. In more open environments where most candidate paths are valid, the relative gains would be smaller, potentially reaching the regime where exhaustive enumeration becomes competitive. This highlights that our framework is most beneficial for challenging geometric scenarios.

\subsubsection{Generalization and Robustness}

The evaluation on diverse training scenarios with randomized building removal, variable heights, and stochastic transmitter-receiver placement suggests reasonable generalization within the class of urban street canyon geometries. However, the framework has important structural limitations that merit acknowledgment.

First, the model was trained exclusively on variations of a single canonical street canyon geometry. The Manhattan evaluation provides critical evidence of this limitation: while the model successfully identifies some principal reflection components across different spatial regions, its generalization to the real-world Manhattan scenes is imperfect. Specifically, the evaluation reveals that the model exhibits non-uniform performance: certain dominant reflection paths are consistently discovered across the coverage map, while other regions---even those dominated by first-order reflections---exhibit strong discrepancies with the ground truth. Moreover, the model fails entirely to discover reflection paths from specific building facades, and some valid paths are only recovered within limited spatial regions rather than everywhere they could occur. This spatial heterogeneity in performance suggests that the model has not fully learned the underlying geometric principles of reflection, but rather has adapted to specific structural patterns present in the training scenarios.

These results indicate that the current model is not yet ready for deployment on substantially different urban morphologies without additional work. The model does not transfer sufficiently to real-world geometries that exhibit different building layouts, street widths, facade complexities, and block organization patterns. Addressing this limitation requires future research. One promising direction is to increase the representational capacity of the model through richer feature encodings or more expressive architectures. Another approach is to directly train on medium-sized scenes (similar to the Manhattan scenes in terms of object count and complexity) from the outset, and subsequently evaluate whether generalization to both smaller and larger scenes improves compared to the current training strategy. This would test whether the current model architecture fundamentally lacks the capacity to represent larger scenes, or whether the training distribution simply does not expose sufficient geometric diversity at that scale.

Second, the framework assumes that the scene geometry is known precisely. In real-world scenarios, site surveys may contain measurement uncertainty or incomplete information about building materials and exact boundaries. The sensitivity of the model to such uncertainties remains unexplored.

Third, our evaluation focuses on the geometric aspect of path finding. However, the contribution of each path in the received signal is influenced by frequency-dependent effects like antenna patterns, polarization, and frequency-selective reflection coefficients, which are handled by the ray tracing engine rather than the learned model. As our model is trained independently of those effects, it could potentially learn sampling strategies that are optimal for geometric validity but suboptimal for signal contribution (e.g., total received power).

\subsubsection{Architectural Choices and Design Trade-offs}

Several design decisions in our neural network architecture deserve brief discussion. The choice of embedding dimension ($d=128$) and the specific network sizes were driven by preliminary experiments, while a more systematic hyperparameter search might reveal improvements. The object encoder, a relatively simple \gls{mlp} applied independently to each facet, does not explicitly capture relationships between adjacent or nearby objects. More sophisticated architectures that employ visibility graphs, graph neural networks, or transformer-based attention mechanisms could better capture multi-scale geometric relationships and potentially improve performance. However, they could also increase computational complexity and reduce the benefits of increased speed.

One important limitation is the fixed number of embeddings used to represent the scene. Although Deep Sets guarantees permutation invariance, it has a known theoretical limitation: it struggles to model complex relational interactions between elements. The aggregation operation compresses the set into a single vector, averaging the features. If the number of objects increases significantly, this averaging may dilute important information about individual objects or their relationships. Because ray tracing relies heavily on the relative positions and orientations of surfaces (e.g., for reflection), this loss of relational structure likely hinders the ability of the model to learn high-order interactions effectively. Future work could explore more expressive set representations that retain richer relational information, including visibility-graph encodings or graph neural networks.

\subsubsection{Broader Impact and Future Directions}

The successful application of the framework to radio coverage mapping suggests its potential extension to other types of ray interactions, such as refraction or diffraction. The key requirement is that the simulated scenes exhibit enough geometric regularity to allow neural networks to learn patterns while still having sufficient randomness to necessitate intelligent sampling. Scattering, by contrast, would require a different action space because it induces a continuum of outgoing directions and is therefore outside the current formulation.

A significant simplification in our evaluation is the assumption that all reflection coefficients are unity, which is unrealistic for real-world propagation. In practice, reflection coefficients are material- and frequency-dependent, typically ranging between 0.3 and 0.7 (or lower), meaning that higher-order reflected paths attenuate dramatically. Consequently, some paths discovered by the model may contribute negligibly to the total received power. Future work should investigate whether incorporating realistic electromagnetic material properties and frequency-dependent reflection coefficients alters the relative importance of discovered paths. This would help determine whether the learned sampling strategy remains effective when paths are weighted by their actual physical contribution, rather than valued equally as geometric solutions.

Future work should address the generalization concerns raised above by: (1) conducting transfer learning experiments with other urban geometries and building morphologies, (2) quantifying uncertainty to evaluate robustness against inaccurate scene descriptions, and (3) optimizing the entire pipeline end-to-end to maximize the final application's accuracy rather than intermediate metrics such as the hit rate. For example, one could investigate whether incorporating additional physics-based features into the input of the model improves the relevance of sampled paths to the final coverage prediction task.

Finally, integrating learned models with classical algorithms deserves deeper investigation. Our framework uses neural networks to augment classical ray tracing techniques. However, the boundary between where learning excels and where classical computation is more appropriate remains uncertain. Hybrid architectures that dynamically select learned or classical components based on input characteristics could be more efficient than the fixed approach taken here.

\section{Conclusion}

In this work, we presented an enhanced machine-learning-assisted framework for point-to-point ray tracing, specifically designed to overcome the computational bottlenecks of exhaustive path searching. By integrating \glspl{gflownet} into the traditional ray tracing pipeline, we successfully transformed the combinatorial challenge of path enumeration into an intelligent, sequential sampling process. Our architectural refinements---namely the successful experience replay buffer, targeted exploratory policy, and physics-based action masking---collectively address the issues of reward sparsity and overfitting that hindered previous iterations of this framework.

Theoretical analysis and experimental results in urban street canyon environments demonstrate that our model maintains the physical interpretability and accuracy of gold-standard ray tracing while achieving linear inference complexity per sample with respect to scene size. Specifically, coverage map predictions show a strong alignment with ground-truth data, though the omission of some higher-order reflection paths and minor near-field instabilities highlight areas for further improvement. In complex scenarios with interaction orders $K \ge 3$, our lightweight, invariant architecture offers substantial speedups and significant memory savings by avoiding the need to store millions of physically invalid path candidates. This makes the framework particularly promising for large-scale digital twins and real-time wireless network optimization where traditional methods hit a \textquote{memory wall.}

Future research efforts to further refine this technology should focus on several key areas. First, explicitly embedding the principle of multipath reciprocity---the fact that a valid ray path can be traversed in reverse---within the neural architecture merits deeper investigation. While data augmentation via reversed \gls{tx}-\gls{rx} paths provides a baseline, implementing specialized loss functions that explicitly penalize discrepancies between forward and reversed path predictions would more robustly integrate this physical constraint into the learning objective. Furthermore, subsequent validation should assess robustness under per-object modifications to ensure geometric diversity. This entails refining performance across varied scenes by analyzing the impact of individual translation, rotation, or scaling of environmental features such as buildings and obstacles. Finally, a promising avenue involves integrating this sampling module into fully differentiable channel optimization pipelines, thereby enabling the simultaneous optimization of transceiver placement and environmental configurations---such as antenna positioning---in high-fidelity simulations.

\section*{Declarations}

\subsection*{Author Contributions}

J.E. wrote the entire manuscript and performed all the simulations and research work activities. E.-M.V., V.D.-E., N.D.C., L.J. and C.O. reviewed the manuscript and participated in research meetings with J.E. to discuss the research work and provide feedback.

\subsection*{Funding}

This work was not supported by any funding.

\subsection*{Data Availability}

All the data (code, benchmarks, tests and a tutorial) is available on a public GitHub repository at \url{https://github.com/jeertmans/sampling-paths}.

\subsection*{Conflict of Interest}

The authors declare that they have no conflicts of interest.

\clearpage
\crefalias{section}{appendix}
\begin{appendices}
  \section{Invariance Properties}\label{appendix:invariance}

We analyze the invariance properties of the coordinate transformation defined in \cref{subsec:geometric_transform}. Specifically, we show that the transformation is invariant under global translation, scaling, and azimuthal rotation around the vertical axis $\boldsymbol{e}_z = [0, 0, 1]^\top$.

\begin{proof}
  Let $f(\boldsymbol{x}_i)$ denote the coordinate transformation. We aim to prove invariance under the transformation $\boldsymbol{x}_i \mapsto \alpha \boldsymbol{Q} \boldsymbol{x}_i + \boldsymbol{b}$, where $\alpha > 0$ is a scaling factor, $\boldsymbol{b} \in \mathbb{R}^3$ is a translation vector, and $\boldsymbol{Q} \in \mathrm{SO}(3)$ is a rotation matrix representing an azimuthal rotation (i.e., $\boldsymbol{Q} \boldsymbol{e}_z = \boldsymbol{e}_z$).

  Let the transformed positions be $\boldsymbol{x}_{\mathrm{\glsxtrshort{tx}}}' = \alpha \boldsymbol{Q} \boldsymbol{x}_{\mathrm{\glsxtrshort{tx}}} + \boldsymbol{b}$, $\boldsymbol{x}_{\mathrm{\glsxtrshort{rx}}}' = \alpha \boldsymbol{Q} \boldsymbol{x}_{\mathrm{\glsxtrshort{rx}}} + \boldsymbol{b}$, and $\boldsymbol{x}_i' = \alpha \boldsymbol{Q} \boldsymbol{x}_i + \boldsymbol{b}$.

  \textbf{1. Translation invariance.}
  The translation vector $\boldsymbol{b}$ cancels out in all difference vectors:
  $\boldsymbol{x}_{\mathrm{\glsxtrshort{rx}}}' - \boldsymbol{x}_{\mathrm{\glsxtrshort{tx}}}' = \alpha \boldsymbol{Q}(\boldsymbol{x}_{\mathrm{\glsxtrshort{rx}}} - \boldsymbol{x}_{\mathrm{\glsxtrshort{tx}}})$ and $\boldsymbol{x}_i' - \boldsymbol{x}_{\mathrm{\glsxtrshort{tx}}}' = \alpha \boldsymbol{Q}(\boldsymbol{x}_i - \boldsymbol{x}_{\mathrm{\glsxtrshort{tx}}})$.
  Thus, the transformed coordinate $\boldsymbol{x}_i'$ is independent of $\boldsymbol{b}$.

  \textbf{2. Scaling invariance.}
  The new scaling factor is $s' = \|\boldsymbol{x}_{\mathrm{\glsxtrshort{rx}}}' - \boldsymbol{x}_{\mathrm{\glsxtrshort{tx}}}'\| = \alpha \|\boldsymbol{Q}(\boldsymbol{x}_{\mathrm{\glsxtrshort{rx}}} - \boldsymbol{x}_{\mathrm{\glsxtrshort{tx}}})\| = \alpha s$.
  In \eqref{eq:transform}, the factor $\alpha$ in the numerator $(\boldsymbol{x}_i' - \boldsymbol{x}_{\mathrm{\glsxtrshort{tx}}}')$ is cancelled by the $\alpha$ in the denominator $s'$.

  \textbf{3. Rotation invariance (Azimuthal).}
  Under the azimuthal rotation $\boldsymbol{Q}$, the longitudinal axis transforms as
  \begin{equation}
    \boldsymbol{w}' = \frac{\boldsymbol{x}_{\mathrm{\glsxtrshort{rx}}}' - \boldsymbol{x}_{\mathrm{\glsxtrshort{tx}}}'}{s'} = \frac{\alpha \boldsymbol{Q} (\boldsymbol{x}_{\mathrm{\glsxtrshort{rx}}} - \boldsymbol{x}_{\mathrm{\glsxtrshort{tx}}})}{\alpha s} = \boldsymbol{Q} \boldsymbol{w}.
  \end{equation}
  For the lateral axis $\boldsymbol{u}$, we use the property that $\boldsymbol{Q}$ is a rotation matrix ($\boldsymbol{Q} \in \mathrm{SO}(3)$), meaning $\boldsymbol{Q}(\boldsymbol{a} \times \boldsymbol{b}) = (\boldsymbol{Q}\boldsymbol{a}) \times (\boldsymbol{Q}\boldsymbol{b})$. Given $\boldsymbol{Q} \boldsymbol{e}_z = \boldsymbol{e}_z$, the vector after rotation is expressed as
  \begin{equation}
    \boldsymbol{u}' = \frac{\boldsymbol{w}' \times \boldsymbol{e}_z}{\|\boldsymbol{w}' \times \boldsymbol{e}_z\|} = \frac{(\boldsymbol{Q} \boldsymbol{w}) \times (\boldsymbol{Q} \boldsymbol{e}_z)}{\|\boldsymbol{Q} (\boldsymbol{w} \times \boldsymbol{e}_z)\|} = \frac{\boldsymbol{Q} (\boldsymbol{w} \times \boldsymbol{e}_z)}{\|\boldsymbol{w} \times \boldsymbol{e}_z\|} = \boldsymbol{Q} \boldsymbol{u}.
  \end{equation}
  Similarly, for the local vertical axis $\boldsymbol{v}$, we have that
  \begin{equation}
    \boldsymbol{v}' = \boldsymbol{w}' \times \boldsymbol{u}' = (\boldsymbol{Q} \boldsymbol{w}) \times (\boldsymbol{Q} \boldsymbol{u}) = \boldsymbol{Q} (\boldsymbol{w} \times \boldsymbol{u}) = \boldsymbol{Q} \boldsymbol{v}.
  \end{equation}
  The new basis matrix is $\boldsymbol{R}' = [\boldsymbol{u}', \boldsymbol{v}', \boldsymbol{w}']^\top = [\boldsymbol{Q}\boldsymbol{u}, \boldsymbol{Q}\boldsymbol{v}, \boldsymbol{Q}\boldsymbol{w}]^\top = \boldsymbol{R} \boldsymbol{Q}^\top$.
  Substituting these into the final transformation gives
  \begin{equation}
    \boldsymbol{x}''_i = \boldsymbol{R}' \left( \frac{\boldsymbol{x}_i' - \boldsymbol{x}_{\mathrm{\glsxtrshort{tx}}}'}{s'} \right) = (\boldsymbol{R} \boldsymbol{Q}^\top) \left( \frac{\alpha \boldsymbol{Q} (\boldsymbol{x}_i - \boldsymbol{x}_{\mathrm{\glsxtrshort{tx}}})}{\alpha s} \right) = \boldsymbol{R} \boldsymbol{Q}^\top \boldsymbol{Q} \left( \frac{\boldsymbol{x}_i - \boldsymbol{x}_{\mathrm{\glsxtrshort{tx}}}}{s} \right).
  \end{equation}
  Since $\boldsymbol{Q}^\top \boldsymbol{Q} = \boldsymbol{I}$, we obtain $\boldsymbol{x}''_i = \boldsymbol{x}'_i$, proving the invariance.
\end{proof}

\end{appendices}

%%===========================================================================================%%
%% If you are submitting to one of the Nature Portfolio journals, using the eJP submission   %%
%% system, please include the references within the manuscript file itself. You may do this  %%
%% by copying the reference list from your .bbl file, paste it into the main manuscript .tex %%
%% file, and delete the associated \verb+\bibliography+ commands.                            %%
%%===========================================================================================%%

\clearpage
\bibliography{biblio}% common bib file
%% if required, the content of .bbl file can be included here once bbl is generated
%%\input sn-article.bbl

\end{document}